%% file: main.tex
\documentclass[sigconf]{acmart}

\usepackage{amsmath,amsfonts} 
\usepackage{algorithmic}
\usepackage{graphicx}
\usepackage{textcomp}
\usepackage{xcolor, colortbl}
\usepackage{times}
\usepackage{soul}
\usepackage{url}
\usepackage{adjustbox}
\usepackage{tabularx}
\usepackage{xspace}
\usepackage{booktabs, caption}
\usepackage{algorithm}
\usepackage{multirow}
\usepackage{array}

\usepackage{bm}
\usepackage[flushleft]{threeparttable}
\usepackage{multirow}
\usepackage{enumitem}

\usepackage[normalem]{ulem}
\usepackage{subcaption}

\newcommand{\cj}[1]{{}}
\newcommand{\cjj}[1]{{#1}}

\newcounter{gaocomm} 
  
\definecolor{blue-violet}{rgb}{0.0, 0.81, 0.90}
\definecolor{mygreen}{rgb}{0.0, 0.5, 0.0}
\definecolor{awesome}{rgb}{1.0, 0.13, 0.32}
\definecolor{bostonuniversityred}{rgb}{0.8, 0.0, 0.0}

\hypersetup{
    colorlinks=true,
    linkcolor=black,
    filecolor=black,
    urlcolor=black,
    citecolor=black,
}

\usepackage{xspace}
\usepackage{adjustbox}

\AtBeginDocument{%
  \providecommand\BibTeX{{%
    \normalfont B\kern-0.5em{\scshape i\kern-0.25em b}\kern-0.8em\TeX}}}

\setcopyright{acmcopyright}
\copyrightyear{2018}
\acmYear{2018}
\acmDOI{XXXXXXX.XXXXXXX}

\acmConference[Conference acronym 'XX]{Make sure to enter the correct
  conference title from your rights confirmation emai}{June 03--05,
  2018}{Woodstock, NY}
\acmPrice{15.00}
\acmISBN{978-1-4503-XXXX-X/18/06}

\begin{document}

\title{SA-MLP: Distilling Graph Knowledge from GNNs into Structure-Aware MLP}

\author{Jie Chen$^{1}$, Shouzhen Chen$^{1}$, Mingyuan Bai$^{1,2}$, Junbin Gao$^{2}$, Junping Zhang$^{1}$, Jian Pu$^{3}$}

\affiliation{
  \institution{$^1$Shanghai Key Lab of Intelligent Information Processing, School of Computer Science, Fudan University, China}
  \institution{$^2$Discipline of Business Analytics, The University of Sydney Business School, The University of Sydney, Australia}
  \institution{$^3$Institute of Science and Technology for Brain-Inspired Intelligence, Fudan University,  China}
  \country{}
}

\affiliation{
  \institution{ $^{1,3}$\{chenj19, chensz19, jpzhang,jianpu\}@fudan.edu.cn, $^{1,2}$yvonne.mingyuanbai@gmail.com, $^{2}$junbin.gao@sydney.edu.au}
  \country{ }
}


\renewcommand{\shortauthors}{Trovato and Tobin, et al.}

\begin{abstract}
    The message-passing mechanism helps Graph Neural Networks (GNNs) achieve remarkable results on various node classification tasks. Nevertheless, the recursive nodes fetching and aggregation in message-passing cause inference latency when deploying GNNs to large-scale graphs. One promising inference acceleration direction is to distill the GNNs into message-passing-free student multi-layer perceptrons (MLPs). However, the MLP student cannot fully learn the structure knowledge due to the lack of structure inputs, which causes inferior performance in the heterophily and inductive scenarios. To address this, we intend to inject structure information into MLP-like students in low-latency and interpretable ways. Specifically, we first design a Structure-Aware MLP (SA-MLP) student that encodes both features and structures without message-passing. Then, we introduce a novel structure-mixing knowledge distillation strategy to enhance the learning ability of MLPs for structure information. Furthermore, we design a latent structure embedding approximation technique with two-stage distillation for inductive scenarios. Extensive experiments on eight benchmark datasets under both transductive and inductive settings show that our SA-MLP can consistently outperform the teacher GNNs, while maintaining faster inference as MLPs. 
    The source code of our work can be found in \url{https://github.com/JC-202/SA-MLP}.
\end{abstract}




\maketitle

\input{sections/Introduction}
\input{sections/Preliminary}
\input{sections/Related_Works}

\input{sections/Method}
\input{sections/Experiments}

\input{sections/Conclusion}

\bibliographystyle{ACM-Reference-Format}
\bibliography{ref}

\input{sec-appendix}
\clearpage

\end{document}

%% file: sections/Introduction.tex
\section{Introduction}

\begin{figure}[t]
    \centering
    \includegraphics[width=1\linewidth]{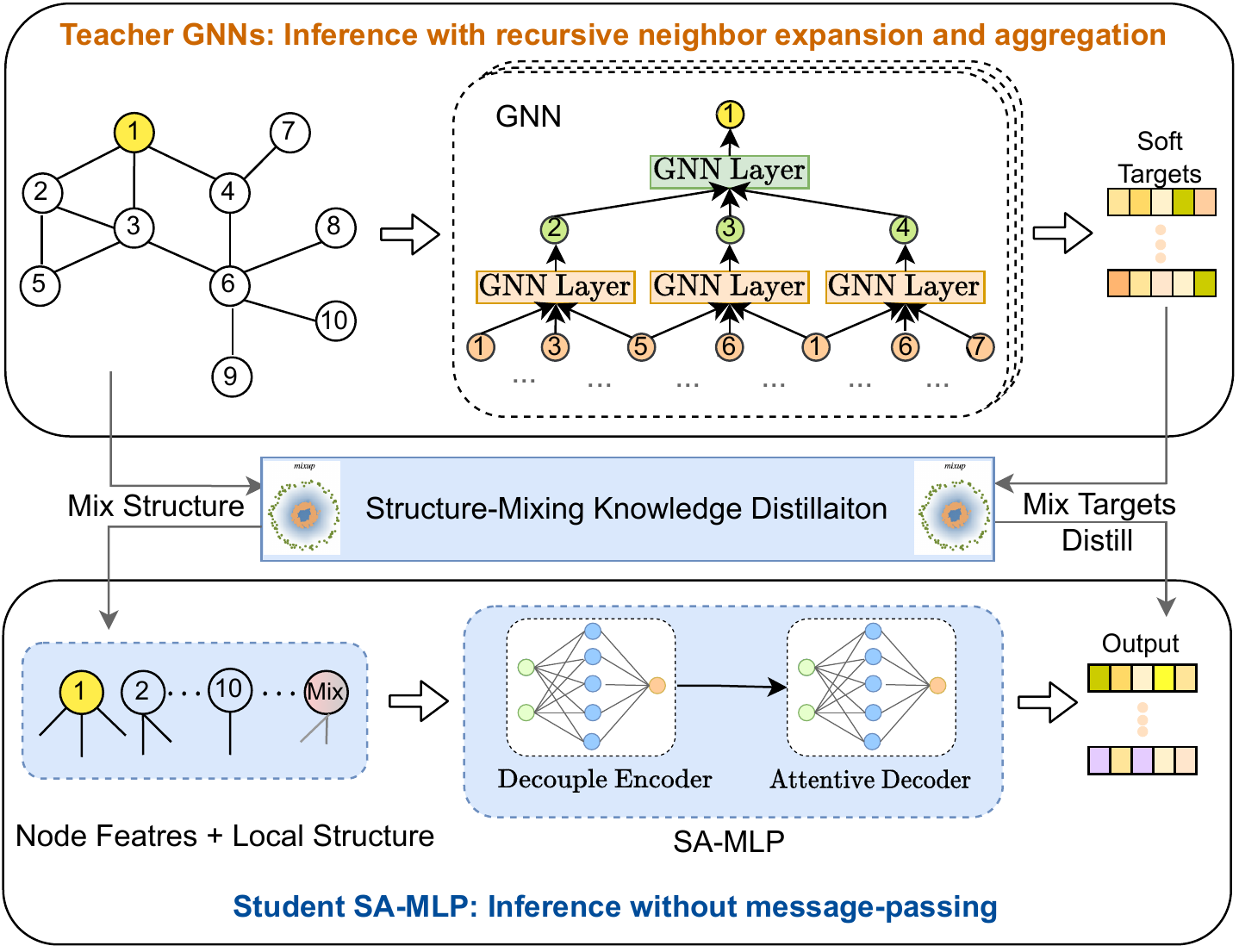}
    \caption{An overview of our distillation framework. A structure-awareness MLP-like student learns from GNNs via a structure-mixing knowledge distillation strategy to achieve substantially faster inference without message-passing.}
    \label{fig:setting1}
\end{figure}

Graph Neural Networks (GNNs)~\cite{GCN,Hamilton2017InductiveRL} have recently emerged as a powerful class of deep learning architectures to analyze graph datasets in diverse domains such as social networks~\cite{sankar2021socialnet}, traffic networks~\cite{wang2020traffic} and recommendation systems~\cite{wang2019knowledge}.
Most GNNs follow a message-passing mechanism~\cite{gilmer2017neural} that extracts graph knowledge by aggregating neighborhood information iteratively to build node representation. However, the number of neighbors for each node would exponentially increase as 
the number of layers increases~\cite{zhang2021graph,yan2020tinygnn}. 
Hence, \cjj{as the yellow node is shown in the upside of Figure~\ref{fig:setting1}}, this recursive neighbor fetching induced by message-passing leads to inference latency, making GNNs hard to deploy for latency-constrained applications that require fast inference, especially for large-scale graphs.

Common inference acceleration methods, such as pruning~\cite{zhou2021accelerating} and quantization~\cite{tailor2020degree,zhao2020learned}, can speed up GNNs to some extent by reducing the Multiplication-and-ACcumulation (MAC) operations.
However, they are still limited by recursive aggregation due to message-passing. Knowledge distillation (KD) is another generic neural network learning paradigm for deployment that transfers knowledge from high-performance but resource-intensive teacher models to resource-efficient students~\cite{hinton2015distilling}. Motivated by the promising results of MLP-like models in computer vision~\cite{melas2021you,liu2022we,Liu2021PayAT}, graph-less neural networks (GLNN)~\cite{zhang2021graph} were proposed to transfer the graph knowledge from GNNs to standard MLP students via KD. 
This idea improves the performance of MLPs on node classification while being easy to deploy in production systems because of discarding the message-passing. 

However, traditional MLPs do not fully understand graph knowledge due to the lack of structure inputs. As in GLNN~\cite{zhang2021graph}, distillation may fail when node labels are highly correlated with structure information, e.g., heterophily datasets. Hence the improvement of distillation is mainly attributed to the strong memory ability of MLPs~\cite{szegedy2013intriguing}.
To better understand the limitations of GLNN, we consider the two scenarios, i.e., transductive (\textit{trans}) and inductive (\textit{ind}), according to information from graphs obtained during the training phase.
In \textit{trans} settings~\cite{GCN}, all node features and graph structures are given in the training time, and student MLPs can overfit (memorize) teachers' outputs on all nodes, which leads to superior performance~\cite{zhang2021graph}. Nevertheless, in the more challenging \textit{ind} scenario~\cite{Hamilton2017InductiveRL} where the test node information is unavailable in the training stage, an MLP without graph dependency has limited generalizability on these test nodes. In this scenario, the structure information is a key clue that binds the training and test nodes to improve generalization.
Furthermore, the standard logit-based KD~\cite{hinton2015distilling}, which merely considers label information on existing nodes, cannot fully transfer the graph knowledge due to the sparsity of the graph structure~\cite{kuramochi2005finding}.

To address the above problems, we intend to inject the structure information into MLPs in a low-latency and interpretable way. To this end, as shown in Figure~\ref{fig:setting1}, we first design a simple yet effective and interpretable Structure-Aware MLP (SA-MLP) student model (Section~\ref{Sec:Method-Student}) to encode both node features and structure information. 
Specifically, the SA-MLP decouples the features and structure information of each node with two encoders, and utilizes an adaptive fusion decoder to generate the final prediction. All the modules of SA-MLP are implemented by MLPs, and the structure inputs are the batch of the sparse adjacency matrix rows. Hence, it can benefit from both mini-batch training and faster inference. 
Second, we propose a novel structure-mixing knowledge distillation (Section~\ref{Sec:Method-KD}) via the mixup~\cite{zhang2018mixup} technique to improve the learning ability of SA-MLP for structure information from GNNs. It generates the virtual mixing structure and teachers' output samples to enhance the distillation density for structure knowledge. 
Compared with standard logit-based distillation, our strategy is more appropriate for SA-MLP due to the reduction of structure sparsity. 
Third, for the \textit{ind} scenario without connection, e.g., some newest users on Twitter who do not interact with any others, we propose an implicit structure embedding approximation technique with a two-stage distillation procedure to enhance the generalization ability (Section~\ref{Sec:Method-Train}). 

We conduct extensive experiments on eight public benchmark datasets under \textit{trans} and \textit{ind} scenarios. The results show that the learned SA-MLP can achieve similar or even better performance than teacher GNNs in all scenarios, with substantially faster inference. Furthermore, we also conduct an in-depth analysis to investigate the compatibility, interpretability, and efficiency of the learned SA-MLP.

To summarize, this work has the following main contributions:
\begin{itemize}
    \item We propose a message-passing free model SA-MLP, which has a low latency inference and a more interpretable prediction process, and naturally preserves the feature/structure information after distillation from GNN.
    \item We design a mixup-based structure-mixing knowledge distillation strategy that improves the performance and structure awareness of SA-MLP via KD. 
    \item For the missing structure scenario, we propose a latent structure embedding approximation technique and a two-stage distillation to enhance the generalization ability.
\end{itemize}


%% file: sections/Preliminary.tex
\section{Preliminary}\label{Sec:3}
\subsection{Notation and Problem Setting}
Consider a graph ${\mathcal{G} = (\mathcal{V}, \mathcal{E})}$, with $N$ nodes and $E$ edges. 
Let ${\mathbf{A}\in \mathbb{R}^{N\times N}}$ be the adjacency matrix, with $\mathbf{A}_{i,j} = 1$ if edge$(i, j) \in \mathcal{E}$, and 0 otherwise. Let ${\mathbf{D}\in \mathbb{R}^{N\times N}}$ be the diagonal degree matrix. 
Each node $v_i$ is given a $d$-dimensional feature representation $\mathbf{x}_i$ and a $c$-dimensional one-hot class label $\mathbf{y}_i$. 
The feature inputs are then formed by $\mathbf{X} \in \mathbb{R}^{N\times d}$, and the labels are represented by $\mathbf{Y} \in \mathbb{R}^{N\times c}$. The labeled and unlabeled node sets are denoted as $\mathcal{V}_L$ and $\mathcal{V}_U$, and we have $\mathcal{V}=\mathcal{V}_L\cup \mathcal{V}_U$.

The task of node classification is to predict the labels $\mathbf{Y}$ by exploiting the nodes' features $\mathbf{X}$ and the graph structure $\mathbf{A}$. 
The goal of our paper is to learn an MLP-like student, such that the learned MLP can achieve similar or even better performance compared with a GNN trained by the same training set, with a much lower computational cost in the inference time. 

\subsection{Graph Neural Networks}
Most existing GNNs follow the message-passing paradigm
which contains node feature transformation and information aggregation from connected neighbors on the graph structure~\cite{gilmer2017neural}.
The general $k$-th layer graph convolution for a node $v_i$ can be formulated as 
\begin{align}
\mathbf{h}_{i}^{(k)}&=f\left(\mathbf{h}_{i}^{(k-1)},\left\{\mathbf{h}_{j}^{(k-1)}: j \in \mathcal{N}(v_i)\right\}\right),
\end{align}
where representation $\mathbf{h}_i$ is updated iteratively in each layer by collecting messages from its neighbors denoted as $\mathcal{N}(v_i)$.
The graph convolution operator $f$ is usually implemented as a weighted sum of nodes' representation according to the adjacent matrix ${\bf A}$ as in GCN~\cite{GCN} and GraphSAGE~\cite{Hamilton2017InductiveRL} or the attention mechanism in GAT~\cite{GAT}. 
However, this recursive expansion and aggregation of neighbors cause inference latency, because the number of neighbors fetching will exponentially increase with increasing layers~\cite{zhang2021graph,yan2020tinygnn}.

The objective function for training GNNs is the cross-entropy of the ground truth labels $\mathbf{Y}$ and the output of the network $\mathbf{\hat{Y}} \in \mathbb{R}^{N \times c}$:
\begin{equation}
    \mathcal{L}_{CE}(\mathbf{\hat{Y}}_{L},\mathbf{Y}_{L}) = -\sum_{i \in \mathcal{V}_L} \sum_{j=1}^{c} \mathbf{Y}_{i j} \ln \mathbf{\hat{Y}}_{i j}.
\end{equation}
\subsection{Transductive and Inductive Scenarios}
\begin{figure}[t]
    \centering
    \includegraphics[width=0.5\textwidth]{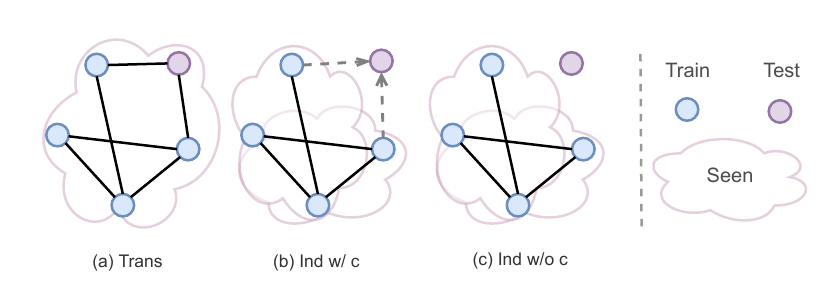}
    \caption{Comparison between transductive and inductive.}
    \label{fig:setting}
\end{figure}

Although the idealistic transductive is the commonly studied setting for node classification, it is incongruous with unseen nodes in real applications. We then consider node classification under three settings to give a broad 
evaluation of models: transductive (\textit{trans}), inductive with connection (\textit{ind w/c}) and inductive without connection (\textit{ind w/o c}), as shown in Figure~\ref{fig:setting}. For \textit{trans}, models can utilize all node features $\mathbf{X}$ and graph $\mathcal{G}$ during training, e.g., to classify unlabeled users in static social networks based on all user features and a small set of labeled users. 
For \textit{ind}, models can not access to the test data during the training stage, which is the same as standard supervised learning. 
However, in the \textit{ind w/c} setting, we may also have access to the structure that connects the training nodes and test nodes. Thus, models can utilize both structure information and node features to predict node labels. It is most related to the ``warmup" scenario in social media networks, where each new user of the system is obligated to pick several existing popular users at the first time. For the \textit{ind w/o c}, test nodes have no connection, aka the ``cold start" of the recommendation system.

Unlike GLNN~\cite{zhang2021graph}, in this work, we carefully consider whether the connection of the test nodes exists and evaluate the performance of teacher GNNs and corresponding MLP-like students in a comprehensive way in these real-world scenarios. Moreover, we test the mixed scenario (\textit{ind w/c} and \textit{ind w/o c}) in Section~\ref{sec:mixed_ind}.

%% file: sections/Related_Works.tex
\section{Related Work}\label{Sec:2}
\subsection{GNNs and Inference Acceleration}
Most GNNs follow the message-passing mechanism~\cite{gilmer2017neural}. For example, GCN~\cite{GCN} aggregates first-order neighbor information according to the Laplacian matrix, GAT~\cite{GAT} employs attention in the aggregation, GraphSAGE~\cite{Hamilton2017InductiveRL} introduces learnable aggregator functions to incorporate local neighborhood,
GCNII~\cite{chen2020simple} introduces residual and initial connections. However, all of these suffer from the inference latency induced by the recursive aggregation. 
Some existing work focuses on speeding up GNN inference from the model compression perspective by pruning GNN parameters~\cite{zhou2021accelerating} and quantizing with low-precision integer arithmetic~\cite{zhao2020learned}, such as Binarized DGCNN~\cite{bahri2021binary} and Degree-quant~\cite{tailor2020degree}.
Nevertheless, 
these approaches can reduce model parameters and MACs operations, 
they are still limited by the neighbor-fetching latency. 
By using contrastive learning to train an MLP, Graph-MLP also makes an attempt to avoid neighbor fetching~\cite{hu2021graph}, but it only considers transductive rather than the more practical inductive setting.
There is also a line of research work for neighbor sampling work~\cite{zou2019layer,chen2018fastgcn} to speed up GNN training, which is complementary to our goal of inference acceleration.

\subsection{Knowledge Distillation for GNNs}
Knowledge distillation~\cite{hinton2015distilling} aims to compress knowledge in a pretrained large teacher model into a compact and fast-to-execute student model. The key idea is to force small student networks to imitate the soft targets generated by the teachers, e.g., minimize the Kullback–Leibler divergence (KL-divergence) between the logit of teacher and student. Existing GNN KD works try to distill large GNNs into smaller GNNs~\cite{Zheng2022ColdBD, Joshi2021OnRK}. For instance, LSP~\cite{yang2020distilling} and TinyGNN~\cite{yan2020tinygnn} conduct KD while preserving local information, GFKD~\cite{Deng2021GraphFreeKD} and DFAD~\cite{Zhuang2022DataFreeAK} achieve graph-level KD via graph generation and adversarial training. Moreover, CPF~\cite{yang2021extract} utilizes KD to learn a label propagation student and enjoy the prior knowledge. However, these methods still require latency-inducing fetching induced by message-passing~\cite{zhang2021graph}. 
To eliminate the message-passing, GLNN~\cite{zhang2021graph} teaches pure MLP student graph knowledge via KD from a teacher GNN.
However, as discussed in the GLNN, the MLP student may fail when structure information is essential, which implies that the MLP does not fully understand structure information. In contrast, we propose to inject structure information into the MLP explicitly while keeping faster inference via a novel structure-aware MLP student and a structure-mixing distillation strategy.

%% file: sections/Method.tex
\section{Proposed Method}\label{Sec:Method}
The key idea of our approach is to make full use of both node feature and structure information for node classification without message-passing. We explicitly inject the structure knowledge into a low-latency message-passing free MLP with a carefully designed student architecture and a novel distillation strategy.

In this section, we first present the Structure-Aware MLP (SA-MLP) which decouples the embedding for node features and structures and merges them by an adaptive late fusion decoder (Section~\ref{Sec:Method-Student}). Then the efficient SA-MLP is used as the student model for knowledge distillation. To further improve the effectiveness of knowledge distillation from GNNs to an SA-MLP student, we  introduce a novel structure-mixing knowledge distillation strategy (Section~\ref{Sec:Method-KD}). Afterwards, we show the
overall training and inference with the latent neighbor embedding approximation technique and the two-stage distillation (Section~\ref{Sec:Method-Train}). 
Finally, we discuss the potential interpretability of the student model and the computational complexity of our framework (Section~\ref{Sec:Method-Discuss}). 

\begin{figure*}[t]
    \centering
    \includegraphics[width=\textwidth]{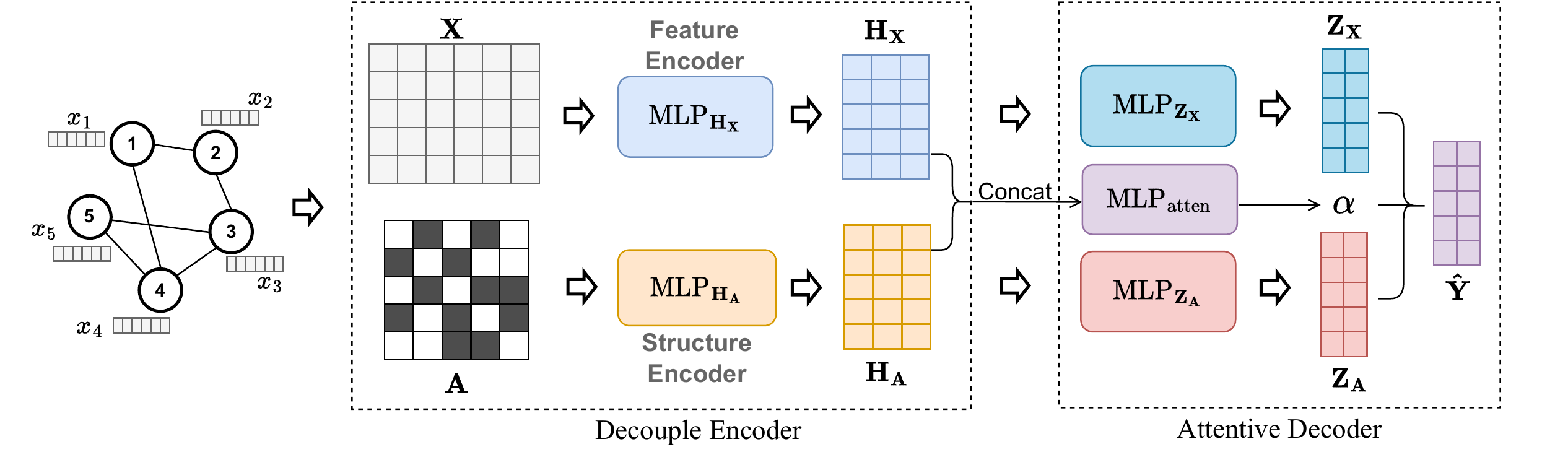}
    \caption{
    Illustration of SA-MLP, which decouples the feature and structure information and fuses them by an attentive late fusion decoder. Different from GNNs, SA-MLP utilizes structure information by treating the sparse adjacency matrix $\mathbf{A}$ columns as features.
    All modules of SA-MLP are pure MLPs to support mini-batch training and fast inference without message-passing.
    }
    \label{fig:overview}
\end{figure*}

\subsection{Structure-Aware MLP Model}\label{Sec:Method-Student}
It is well-known that the required information for node classification contains two parts, i.e., node feature information and structure information~\cite{GCN,zhu2021graph}. 
The importance of each part varies for different nodes and datasets.
To effectively encode both node and structure information and take the unknown contribution of each part into account, as shown in Figure~\ref{fig:overview}, we present the proposed Structure-Aware MLP (SA-MLP) model with the corresponding feature encoder, the structure encoder, and the attentive decoder. 
For inference efficiency, all these modules are implemented with MLPs to extract and fuse the feature and structure information. The proposed SA-MLP model has enough capacity to learn from training data and/or teacher GNNs with a more interpretable and efficient prediction mechanism.

\subsubsection{Decouple Encoder}
GNNs utilize a message-passing mechanism to capture features and structure information simultaneously. In our case, for the feature encoder, we simply utilize an MLP to transform the raw features of nodes into feature embedding that contains self-information. 
However, for the structure encoder, the key question is how to capture the structure information with minimal inference cost without interaction between nodes via message-passing. Inspired by FastText~\cite{joulin2016fasttext} in natural language processing, which represents the sentence as the summation embedding of each sequentially connected word (line-structured), we treat each row of the structure matrix ${\bf A}$ (the connected neighbor of each node in the local structure) as a bag of nodes. 
Note that we remove the self-loop, i.e., the diagonal of ${\bf A}$, to enhance the inductive ability.
Hence, we can feed the sparse row of the adjacency matrix into an MLP to capture the structure information. More precisely, the structure and feature embedding are encoded individually by
\begin{align}
\mathbf{H_A} = \operatorname{MLP_{\mathbf{H_A}}}(\mathbf{{A}}),\quad \mathbf{H_X} = \operatorname{MLP_{\mathbf{H_X}}}(\mathbf{X}),
\end{align}
where the $\operatorname{MLP_{\mathbf{H_X}}}$ and $\operatorname{MLP_{\mathbf{H_A}}}$ encoders can be set as one linear layer to efficiently generate the features embedding $\mathbf{H_X} \in \mathbb{R}^{n \times d}$ and the structure embedding $\mathbf{H_A} \in \mathbb{R}^{n \times d}$ of nodes. 

The structure encoder $\operatorname{MLP_{\mathbf{H_A}}}$ can be regarded as learning the structure positional embedding of each node and capturing the current node local structure information by summarizing all connected nodes' structure positional embedding. This process behavior is also similar to the usage of MLPs to learn user and item embeddings in Recommendation Systems (RSs) for large-scale industrial production~\cite{he2017neural,barkan2016item2vec}.
Similar to the embedding system in RSs, moreover, we may retrain the structure positional embedding to assist future online learning by regularly collecting new edges in the dynamic social networks.
Another advantage of the proposed encoder is that it allows minibatch training and inference, because SA-MLP utilizes graph information solely by defining the adjacency matrix $\mathbf{A}$ columns as features.
\cjj{Additionally, as $\mathbf{A}$ is typically very sparse for real-world networks, the separating of $\mathbf{X}$ and $\mathbf{A}$ enables a sparse-dense matrix product to calculate the mapping of $\operatorname{MLP_{\mathbf{H_A}}}$ on $\mathbf{A}$, substantially increasing efficiency.}
Hence, this design is more scalable to large-scale graphs than vanilla GNNs, which need full-batch message-passing.

\subsubsection{Attentive Decoder}
After encoding features and structure information, we propose an adaptive decoder to generate the final prediction via an interpretable late fusion mechanism. We first apply two MLPs, $\operatorname{MLP_{\mathbf{Z_A}}}$ and $\operatorname{MLP_{\mathbf{Z_X}}}$, to generate the output $\mathbf{Z_A} \in \mathbb{R}^{n \times c}$ and $\mathbf{Z_X} \in \mathbb{R}^{n \times c}$, respectively. 
\begin{align}
\mathbf{Z_A} = \operatorname{MLP_{\mathbf{Z_A}}}(\mathbf{H_A}),\quad \mathbf{Z_X} = \operatorname{MLP_{\mathbf{Z_X}}}(\mathbf{H_X}).
\end{align}
Then, a gating network $\operatorname{MLP}_\text{atten}$ is used to adaptively fuse these two outputs according to the corresponding structure and feature embedding of each node. The gating network is implemented as a linear layer activated by a sigmoid function $\sigma$:
\begin{align}
\bm{\alpha} &= \sigma(\mathbf{W}[\mathbf{H_{A}}||\mathbf{H_{X}}]+b),\\
\hat{\mathbf{{Y}}} &= \operatorname{softmax}((1-\bm{\alpha})  \cdot \mathbf{Z_X} + \bm{\alpha} \cdot \mathbf{Z_A}),
\end{align}
where $\mathbf{W} \in \mathbb{R}^{2d \times 1}$ and $b \in \mathbb{R}$ are trainable parameters. 
The gating network learns to balance two outputs of each node according to the weighting score $\bm{\alpha} \in \mathbb{R}^{N \times 1}$, which also provides explainability for the combination ratio of structure and node features.

Besides the explainable ability, another reason for us to adopt late fusion is its flexibility. When one modality information (structure/features) is missed, we can skip the fusion stage and support the prediction. For instance, in a social network, on the one hand, one lazy user may only follow other users without providing features (self-information). We can support the prediction without features via the downstream structure path of the SA-MLP.
On the other hand, a new user may provide some self-information but not follow others, i.e., without structure-information. As a special case, if $\bm{\alpha}=0$, our SA-MLP also degenerates to GLNN~\cite{zhu2021graph}, which merely takes node features for prediction.  

\subsection{Structure-Mixing Knowledge Distillation }\label{Sec:Method-KD}
Although the SA-MLP can efficiently capture features and structure information, discarding the message-passing between nodes still causes suboptimal performance. To allow the SA-MLP to enjoy both the efficiency of MLP and the accuracy of GNNs, such as GLNN~\cite{zhang2021graph}, we also conduct the cross-model KD from GNNs to SA-MLP. 
Since the outputs of GNNs are considered to include structure information~\cite{yang2021extract,yan2020tinygnn}, the logit-based KD~\cite{hinton2015distilling} has been widely utilized to extract graph knowledge from GNNs.
However, in our case, the standard KD may not be sufficient to help SA-MLP learn meaningful structure embedding due to the sparsity of the graph structure~\cite{kuramochi2005finding}.
For instance, the structure embedding of nodes with only a few connections may not be sufficiently optimized.
Hence, to enhance the awareness of structure, we introduce a novel structure-mixing knowledge distillation strategy.

Inspired by the \textit{mixup} data augmentation strategy in computer vision~\cite{zhang2018mixup}, which generates a virtual vicinal distribution to enhance the generalization via a linear combination of paired inputs $\mathbf{X}$ and labels $\mathbf{Y}$, we design a structure-mixup variant to reduce the sparsity of the structure. It simultaneously mixes features $\mathbf{X}$, structure $\mathbf{A}$, and teacher's output $\mathbf{Y}^t$ to generate virtual distillation samples as follows.

\begin{align}
\text{Structure-Mixup}&\left\{
\begin{aligned}
\lambda &\sim Beta(\eta,\eta) \\ 
\tilde{\mathbf{X}} &= \lambda \mathbf{X} +(1-\lambda)\mathbf{X}_{*,:}\\
\mathbf{\tilde{A}} &= \lambda \mathbf{A} +(1-\lambda)\mathbf{A}_{*,:} \\
\tilde{\mathbf{Y}}^t &= \lambda \mathbf{Y}^t +(1-\lambda)\mathbf{Y}_{*,:}^t
\end{aligned}
\right.
\end{align}
where the hyper-parameter $\eta$ controls the strength of interpolation, and we set it to 0.2 for most experiments. The subscript $_*$ means the index of the corresponding batch sample pair after random shuffling for linear combination, e.g., the row index of nodes from [1,2,...,n] to [5,n-1,...,2] after shuffling.
Then, we forward SA-MLP twice to generate the standard student output $\mathbf{Y}^s$ and mixing output $\mathbf{\tilde{Y}}^s$:
\begin{align}\label{eq:dis}
\mathbf{Y}^s = \operatorname{SA-MLP}(\mathbf{X}, \mathbf{A}),\quad
\tilde{\mathbf{Y}}^s = \operatorname{SA-MLP}(\tilde{\mathbf{X}}, \tilde{\mathbf{A}}).
\end{align}
The \textit{mixup} on graphs is regarded as challenging due to the irregularity and connectivity, and existing \textit{mixup} methods for GNNs aim to mix hidden embedding~\cite{Wang2021MixupFN,Han2022GMixupGD}. However, in our case, our SA-MLP can naturally process the structured agency mixing matrix $\tilde{\mathbf{A}}$. Besides, a recent study in computer vision shows that the \textit{mixup} can enhance the function matching property of KD~\cite{beyer2022knowledge}.
The mixing pair of the structure $\mathbf{\tilde{A}}$ and the teacher's output $\mathbf{\tilde{Y}}^t$, which contains the hybrid graph knowledge samples, can enhance the density of distillation for structure knowledge. 
To the best of our knowledge, we are the first to introduce the \textit{mixup} on the original graph structure of the KD in GNNs, and the structure-mixing strategy is specifically designed for our SA-MLP.
The overall distillation objective is:
\begin{align}\label{eq:dis1}
    \mathcal{L}_{Dis}(\mathbf{Y}^s,\mathbf{Y}^t) &= \sum\nolimits_{v\in{\mathcal V}}^{} (\delta \operatorname{KL}(\mathbf{y}_v^s, \mathbf{y}_v^t)+ (1-\delta)\operatorname{KL}(\mathbf{\tilde{y}}_v^s, \mathbf{\tilde{y}}_v^t))
\end{align}
where KL means the KL-divergence,
and $\delta$ is a weight parameter balancing the standard logit-based distillation $\operatorname{KL}(\mathbf{y}_v^s, \mathbf{y}_v^t)$ and our structure-mixing distillation $\operatorname{KL}(\mathbf{\tilde{y}}_v^s, \mathbf{\tilde{y}}_v^t)$.

After distillation, the SA-MLP is optimized to exploit the structure knowledge and behavior as well as GNNs. Unlike the GLNN~\cite{zhang2021graph} without structure input and structure-mixing distillation, our SA-MLP enjoys the strengths of both structure information and the fast inference speed.

\subsection{Overall Training and Inference}\label{Sec:Method-Train}
In this section, we describe the overall training and inference process of transductive and inductive settings for SA-MLP.  

\subsubsection{Transductive:} In the transductive setting, the model can observe the structure and features of all nodes during training. Hence, the training objective contains the cross-entropy loss with the ground-truth label on training nodes and the distillation loss with the output of teacher GNNs on total nodes. The total objective is:
\begin{align}\label{eq:trans-obj}
\mathcal{L} = (1-\lambda) \mathcal{L}_{CE}(\mathbf{{Y}}^s_{L},\mathbf{{Y}}_{L})+ \lambda \mathcal{L}_{Dis}(\mathbf{{Y}}^s,\mathbf{{Y}}^{t}).
\end{align}
However, the transductive setting may not be sufficient to evaluate the graph knowledge learning ability, since MLP-like models may memorize all the outputs of teacher GNNs.

\subsubsection{Inductive with Connection:}
In the inductive setting, the model can only observe the structure and features of training set nodes during training.
Hence the total objective only involves the training nodes:
\begin{align}\label{eq:weak-ind-obj}
\mathcal{L} = (1-\lambda) \mathcal{L}_{CE}(\mathbf{{Y}}^s_{L},\mathbf{Y}_{L})+ \lambda \mathcal{L}_{Dis}(\mathbf{{Y}}^s_{L},\mathbf{{Y}}^{t}_{L}).
\end{align}
After training, in the \textit{ind w/c}, the SA-MLP can utilize the connection from training nodes to the newest nodes to infer labels, which means the awareness of structure knowledge that differs from GLNN.

\subsubsection{Inductive without Connection:}
In the \textit{ind w/o c} setting, the newest node is totally isolated and contains no structure information, e.g., the new user in Twitter is too lazy to interact with others. 
Lacking the structure information causes inconsistencies between training with connections and inference without connections, which may jeopardize the models' generalization on the newest nodes.
To solve this problem, we introduce how the SA-MLP is aware of these newest nodes' latent structure information to enhance generalization ability.


\noindent
\textbf{Infer with latent structure embedding approximation
:}\label{Sec:Method-Latent}
One popular solution to the structure missing problem is to learn the potential connection based on the existing connection paradigm and node features, also called graph structure learning~\cite{liu2022towards,zhu2021deep}. Its underlying assumption is that similar nodes may share a similar neighbor so that we can establish the latent relationships from the node features $\mathbf{X}$ to the latent structure $\mathbf{A}'$. However, the search space of connection prediction for each new node from existing nodes is $O(N)$ which causes the inferior inference speed. To address this problem, we propose to directly approximate the latent structure embedding $\mathbf{H'_A}$ based on $\mathbf{X}$ for our SA-MLP rather than structure $\mathbf{A}'$. Specifically, following the universal approximation theorem~\cite{hornik1989multilayer}, we apply another $\operatorname{MLP-2}_{\mathbf{H_A}}$ to replace $\operatorname{MLP}_{\mathbf{H_A}}$ in the encoder and capture the latent relation from each node feature to the latent structure embedding:
\begin{align}
    \mathbf{{H}_A} = \operatorname{MLP}_{\mathbf{H_A}}(\mathbf{A}) \to \mathbf{{H}'_A} = \operatorname{MLP-2}_{\mathbf{H_A}}(\mathbf{X}).
\end{align}


\noindent
\textbf{Train with two-stage distillation:} In order to guide the MLP to learn meaningful structure embedding $\mathbf{{H}'_A}$, we utilize the two-stage distillation. In the first stage, train and distill a standard SA-MLP with Eq~\eqref{eq:weak-ind-obj}. It can help the SA-MLP sufficiently learn ``warmup'' parameters and meaningful structure embedding from the existing features and connections. Then, in the second stage, we freeze the parameters of the SA-MLP except for the new $\operatorname{MLP-2}_{\mathbf{H_A}}$, which can preserve the compatibility of the \textit{ind w/c} scenario, e.g., using $\operatorname{MLP-2}_{\mathbf{H_A}}$ for nodes without connections and $\operatorname{MLP}_{\mathbf{H_A}}$  for nodes with connections.
To optimize $\operatorname{MLP-2}_{\mathbf{H_A}}$, a straightforward way is to force the $\operatorname{MLP-2}_{\mathbf{H_A}}$ to generate similar structure embeddings of training nodes as $\operatorname{MLP}_{\mathbf{H_A}}$ in the previous stage, which can establish the underlying mapping from features to structure embedding. However, we found that simply applying distillation with Eq~\eqref{eq:weak-ind-obj} again can achieve good enough results. The reason may be that the two stages of distillation share the same objective and achieve consistency matching while maintaining the distillation process's simplicity.

\subsection{Discussion on Interpretability and Complexity}\label{Sec:Method-Discuss}
After KD, SA-MLP predicts the label of a specific node $v$ as a linear combination between the two streams, i.e., predictions of structure and feature information. The balance score $\alpha_v$ as an element of $\bm{\alpha}$ indicates whether structure or features is more important for node $v'$ prediction. Therefore, the learned SA-MLP has better interpretability than GNN teachers and pure MLP.

The time complexity of SA-MLP in each full-batch forward pass is $O(dE + Nd^2K)$, in which $d$ is the hidden dimension, $N$ 
is the number of nodes, $E$ is the number of edges, and $K$ is the number of layers. The cost is $O(dE)$ for the first linear mapping of $\mathbf{A}$ and $O(d^2)$ for each MLP. In fact, the operations of mapping $\mathbf{A}$ can be easily implemented in the sparse matrix form, which results in high time efficiency.
While message-passing based GNNs have to propagate features using the adjacency in each layer, their complexity is usually $O(dKE + Nd^2K)$, and the $O(dKE)$ term makes it difficult to deploy to large-scale graphs~\cite{yan2020tinygnn}. 

%% file: sections/Experiments.tex
\section{Experiments}\label{Sec:6}
\subsection{Experimental Setup}
\subsubsection{Datasets}
To evaluate the performance of the proposed SA-MLP, we consider eight public benchmark datasets, and the statistics are summarized in Table~\ref{tab:exp-dataset}, including three citation datasets~\cite{sen2008collective} (Cora, Citeseer, Pubmed), two larger OGB datasets~\cite{hu2020open} (Arxiv, Products), and three heterophily datasets~\cite{pei2019geom,lim2021large} that structure information is important (Chameleon, Squirrel, Arxiv-year). We used the standard public splits of OGB datasets, and ten frequently used fully supervised splits (48\%/32\%/20\% of nodes per class for train/validation/test) provided by~\cite{pei2019geom} of other datasets for a fair comparison and reproduction. 
Note that, for citation datasets, these splits reduce randomness and the possibility of overfitting~\cite{zhu2020beyond}, which are stricter than the random splits used in GLNN~\cite{zhang2021graph}. More details can be found in the Appendix.

\begin{table}[!t]
	\caption{Statistics of the datasets}
	\label{tab:exp-dataset}
	\begin{tabular}{c *{4}{r}}
		\toprule
		\textbf{Datasets} & \textbf{\#Nodes} & \textbf{\#Edges} & \textbf{\#Features} & \textbf{\#Classes} \\
		\midrule
		Cora 	& 2,708  & 5,429  & 1,433   & 7  \\
		Citeseer & 3,327 & 4,732 & 3,703 & 6 \\
		Pubmed & 19,717 & 44,324 & 500 & 3\\
		Arxiv & 169,343 & 1,166,243 & 128 & 40\\
		Products & 2,449,029 & 61,859,140 & 100 & 47\\
		Chameleon & 2,277 & 36,101 & 2,325 & 5\\
		Squirrel & 5,201 & 217,073 & 2,089 & 5\\
		Arxiv-year & 169,343 & 1,166,243 & 128 & 5\\
		\bottomrule
	\end{tabular}
\end{table}


\subsubsection{Transductive and Inductive Setting}
For the transductive (\textit{trans}), we use all node features and structures for training and distillation. For the inductive (\textit{ind}), we hold out all test and validation nodes ($\mathcal{V}_U$) with their connections when training. For every split, we extract graph $\mathcal{G}$ to the subgraph $\mathcal{G}_L$ that only contains nodes $\mathcal{V}_L$ with corresponding edges, and the subgraph $\mathcal{G}_{ind}$ including $\mathcal{G}_L$ plus such edges from $\mathcal{V}_L$ to $\mathcal{V}_U$. Concretely, the input/output of all settings are:

\begin{itemize}
    \item \textit{trans}: train on ($\mathcal{G}$, $\mathbf{X}$, $\mathbf{Y}_L$); evaluate on ($\mathcal{G}$, $\mathbf{X}_U$, $\mathbf{Y}_U$), KD for all nodes $\mathcal{V}$.
    \item \textit{ind w/c}: train on ($\mathcal{G}_L$, $\mathbf{X}_L$, $\mathbf{Y}_L$); evaluate on ($\mathcal{G}_{ind}$, $\mathbf{X}_U$, $\mathbf{Y}_U$), KD for $\mathcal{V}_{L}$.
    \item \textit{ind w/o c}: train on ($\mathcal{G}_L$, $\mathbf{X}_L$, $\mathbf{Y}_L$); evaluate on ($\mathbf{X}_U$, $\mathbf{Y}_U$), KD for $\mathcal{V}_{L}$.
\end{itemize}

\subsubsection{Baselines and Training Details}
In the following experiments, as~\cite{zhang2021graph}, we also use GraphSAGE~\cite{Hamilton2017InductiveRL} as our basic teacher model to investigate the learning ability of the proposed SA-MLP from GNNs. 
Moreover, for the heterophily datasets, we apply residual connections to improve the performance of GraphSAGE.
Following the standard setting~\cite{hu2020open, bo2021beyond}, we fix the hidden dimension of SA-MLP as 128 for the OGB dataset (Arxiv and Products) and 64 for others. 
We use Adam~\cite{kingma2014adam} for optimization, LayerNorm~\cite{Ba2016LayerN}, and tune other hyper-parameters (learning rate, weight decay, etc.) via validation sets of each dataset. 
We report results from previous works with the same experimental setup if available. If the results were not previously reported, we conducted a hyper-parameter search based on the official codes.
More detail can be found in the Appendix. 

\subsection{Overall Performance} 
\begin{table*}[htbp]
    \centering
    \caption{\textmd{Experiment results in node classification for the \textbf{\underline{transductive setting}}. We report the mean test accuracy (\%) and standard deviation over ten runs for each dataset. $\bigtriangleup$ represents the improvement of $\text{SA-MLP}^{KD}$, i.e., 
    $\bigtriangleup_{GLNN} \ge 0$ indicates $\text{SA-MLP}^{KD}$ outperforms GLNN; }}
    \begin{threeparttable}
    \renewcommand\tabcolsep{5pt}
    \renewcommand\arraystretch{1.0}
    \begin{tabular}{l|lllll|lll}
        \toprule[1.2pt]
            Dataset &   SAGE      & MLP      & GLNN                & SA-MLP  &  $\text{SA-MLP}^{KD}$   & $\bigtriangleup_{SA-MLP}$ & $\bigtriangleup_{GLNN}$ &  $\bigtriangleup_{GNN}$       \\

         \midrule
Cora & 86.14$\pm$0.74 & 74.75$\pm$2.22 & 86.21$\pm$1.42 & 76.52$\pm$2.56 & $\textbf{86.30}$ $\pm$1.04 & 9.78(12.78\%) & 0.09(0.10\%) & 0.16(0.19\%) \\ 
Citeseer & 75.13$\pm$2.28 & 72.41$\pm$2.18 & 76.15$\pm$2.19 & 71.78$\pm$2.01 & $\textbf{76.37}$ $\pm$1.57 & 4.59(6.39\%) & 0.22(0.29\%) & 1.24(1.65\%) \\ 
Pubmed & 89.17$\pm$0.46 & 86.65$\pm$0.35 & 89.32$\pm$0.43 & 87.05$\pm$0.59 & $\textbf{89.72}$ $\pm$0.30 & 2.67(3.07\%) & 0.40(0.45\%) & 0.55(0.62\%) \\ 
Arxiv & 70.92$\pm$0.17 & 56.05$\pm$0.46 & 63.46$\pm$0.45 & 63.48$\pm$0.46 & $\textbf{71.54}$ $\pm$0.19 & 8.06(12.70\%) & 8.08(12.73\%) & 0.62(0.87\%) \\ 
Product & 78.61$\pm$0.49 & 62.47$\pm$0.10 & 68.86$\pm$0.46 & 75.49$\pm$0.25 & $\textbf{79.02}$ $\pm$0.15 & 3.53(4.68\%) & 10.16(14.75\%) & 0.41(0.52\%) \\ 
Chameleon & 71.38$\pm$1.76 & 46.36$\pm$2.52 & 67.98$\pm$1.71 & 61.91$\pm$2.06 & $\textbf{71.66}$ $\pm$1.54 & 9.75(15.75\%) & 3.68(5.41\%) & 0.28(0.39\%) \\ 
Squirrel & 62.51$\pm$2.01 & 29.68$\pm$1.81 & 62.23$\pm$1.87 & 60.53$\pm$2.79 & $\textbf{65.40}$ $\pm$2.27 & 4.87(8.05\%) & 3.17(5.09\%) & 2.89(4.62\%) \\ 
Arxiv-year & 51.85$\pm$0.22 & 36.71$\pm$0.21 & 46.22$\pm$0.20 & 50.97$\pm$0.26 & $\textbf{53.31}$ $\pm$0.17 & 2.34(4.59\%) & 7.09(15.34\%) & 1.46(2.82\%) \\

        \bottomrule[1.2pt]
    \end{tabular}
     \begin{tablenotes}
        \footnotesize
        \item[1] Results are from our reproduction with the authors’ public released code, as they didn't report the results in part of these splits or datasets. 
    \end{tablenotes}
    \end{threeparttable}
    \label{tab:node_trans}
\end{table*}

\begin{table*}[htbp]
    \centering
    \caption{\textmd{Experiment results in node classification for the \textbf{\underline{inductive with connection setting}}. }}
    \begin{threeparttable}
    \renewcommand\tabcolsep{4.5pt}
    \renewcommand\arraystretch{1}    \begin{tabular}{l|lllll|lll}
        \toprule[1.2pt]
            Dataset &   SAGE      & MLP      & GLNN                & SA-MLP  &  $\text{SA-MLP}^{KD}$   & $\bigtriangleup_{SA-MLP}$ & $\bigtriangleup_{GLNN}$ &  $\bigtriangleup_{GNN}$       \\

         \midrule
Cora & 80.78$\pm$2.44 & 74.75$\pm$2.22 & 74.98$\pm$1.84 & 73.92$\pm$1.86 & $\textbf{81.24}$ $\pm$2.37 & 7.32(9.90\%) & 6.26(8.35\%) & 0.46(0.57\%) \\ 
Citeseer & 73.24$\pm$1.73 & 72.41$\pm$2.18 & 72.55$\pm$1.79 & 72.59$\pm$1.91 & $\textbf{73.42}$ $\pm$1.30 & 0.83(1.14\%) & 0.87(1.20\%) & 0.18(0.25\%) \\ 
Pubmed & 87.98$\pm$0.66 & 86.65$\pm$0.35 & 88.25$\pm$0.43 & 87.03$\pm$0.55 & $\textbf{88.73}$ $\pm$0.54 & 1.70(1.95\%) & 0.48(0.54\%) & 0.75(0.85\%) \\ 
Arxiv & 67.69$\pm$0.24 & 56.05$\pm$0.46 & 56.79$\pm$0.81 & 63.69$\pm$0.19 & $\textbf{68.01}$ $\pm$0.24 & 4.32(6.78\%) & 11.22(19.76\%) & 0.32(0.47\%) \\ 
Product & 65.55$\pm$0.88 & 62.47$\pm$0.10 & 62.45$\pm$0.34 & 65.19$\pm$0.11 & $\textbf{67.46}$ $\pm$0.36 & 2.27(3.48\%) & 5.01(8.02\%) & 1.91(2.91\%) \\ 
Chameleon & 63.73$\pm$1.58 & 46.36$\pm$2.52 & 46.81$\pm$2.12 & 59.23$\pm$2.32 & $\textbf{63.86}$ $\pm$4.61 & 4.63(7.82\%) & 17.05(36.42\%) & 0.13(0.20\%) \\ 
Squirrel & 58.55$\pm$1.47 & 29.68$\pm$1.81 & 30.19$\pm$1.95 & 53.01$\pm$2.87 & $\textbf{64.25}$ $\pm$1.80 & 11.24(21.20\%) & 34.06(112.82\%) & 5.70(9.74\%) \\ 
Arxiv-year & 48.42$\pm$0.46 & 36.71$\pm$0.21 & 36.73$\pm$1.25 & 48.12$\pm$0.30 & $\textbf{49.55}$ $\pm$0.33 & 1.43(2.97\%) & 12.82(34.90\%) & 1.13(2.33\%) \\

        \bottomrule[1.2pt]
    \end{tabular}

    \end{threeparttable}
    \label{tab:node_weak_ind}
\end{table*}

\begin{table*}[htbp]
    \centering
    \caption{\textmd{Experiment results in node classification for the \textbf{\underline{inductive without connection setting}}. $\text{SA-MLP}^{KD^{1}}$ and $\text{SA-MLP}^{KD^{2}}$ indicate the 1st-stage distillation and 2nd-stage distillation, respectively}}
    \begin{threeparttable}
    \renewcommand\tabcolsep{2.5pt}
    \renewcommand\arraystretch{1}
    \begin{tabular}{l|llllll|lll}
        \toprule[1.2pt]
            Dataset &   SAGE      & MLP      & GLNN                & SA-MLP & $\text{SA-MLP}^{KD^{1}}$ & $\text{SA-MLP}^{KD^{2}}$  & $\bigtriangleup_{SA-MLP}$ & $\bigtriangleup_{GLNN}$ &  $\bigtriangleup_{GNN}$       \\

         \midrule

Cora & 73.70$\pm$3.00 & 74.75$\pm$2.22 & 74.65$\pm$2.08 & 72.58$\pm$2.51 & 73.62$\pm$1.83 & $\textbf{74.89}$ $\pm$1.68 & 2.31(3.18\%) & 0.24(0.32\%) & 1.19(1.61\%) \\ 
Citeseer & 72.52$\pm$2.56 & 72.41$\pm$2.18 & 72.75$\pm$1.98 & 71.44$\pm$1.91 & 72.88$\pm$1.95 & $\textbf{73.03}$ $\pm$1.86 & 1.59(2.23\%) & 0.28(0.38\%) & 0.51(0.70\%) \\ 
Pubmed & 86.89$\pm$0.35 & 86.65$\pm$0.35 & 87.78$\pm$0.44 & 87.03$\pm$0.61 & 87.98$\pm$0.44 & $\textbf{88.08}$ $\pm$0.35 & 1.05(1.21\%) & 0.30(0.34\%) & 1.19(1.37\%) \\ 
Arxiv & 51.43$\pm$0.15 & 56.05$\pm$0.46 & 56.22$\pm$0.34 & 54.62$\pm$0.48 & 56.24 $\pm$0.22 & $\textbf{56.24}$$\pm$0.22 & 1.62(2.97\%) & 0.02(0.04\%) & 4.81(9.35\%) \\ 
Product & 55.75$\pm$0.28 & 62.47$\pm$0.10 & 62.44$\pm$0.34 & 59.65$\pm$0.15 & 60.21$\pm$0.19 & $\textbf{62.55}$ $\pm$0.32 & 2.90(4.86\%) & 0.11(0.18\%) & 6.80(12.20\%) \\ 
Chameleon & 46.23$\pm$1.82 & 46.36$\pm$2.52 & 46.78$\pm$2.41 & 43.49$\pm$4.61 & 44.93$\pm$3.67 & $\textbf{47.02}$ $\pm$3.53 & 3.53(8.12\%) & 0.24(0.51\%) & 0.79(1.71\%) \\ 
Squirrel & 28.18$\pm$1.25 & 29.68$\pm$1.81 & 30.08$\pm$1.73 & 30.05$\pm$1.49 & 30.83$\pm$3.10 & $\textbf{31.26}$ $\pm$2.26 & 1.21(4.03\%) & 1.18(3.92\%) & 3.08(10.93\%) \\ 
Arxiv-year & 34.14$\pm$0.37 & 36.71$\pm$0.21 & 36.73$\pm$0.86 & 29.81$\pm$2.01 & 33.27$\pm$1.80 & $\textbf{36.75}$ $\pm$1.62 & 6.94(23.28\%) & 0.02(0.05\%) & 2.61(7.64\%) \\

        \bottomrule[1.2pt]
    \end{tabular}
    \end{threeparttable}
    \label{tab:node_strong_ind}
\end{table*}

Experimental results on eight datasets over three scenarios with teacher GNN, student MLP, GLNN, student SA-MLP, and student $\text{SA-MLP}^{KD}$ that distilled via structure-mixing KD are presented in Table~\ref{tab:node_trans},~\ref{tab:node_weak_ind} and ~\ref{tab:node_strong_ind}. In a nutshell, due to the structure awareness, our $\text{SA-MLP}^{KD}$ consistently achieves the best performance in all settings, and we further make the following observations:

\subsubsection{Transductive}
As shown in Table~\ref{tab:node_trans}, we first notice that although SA-MLP outperforms MLP due to the structure awareness, it is still inferior to GNNs. Then, after KD, all $\text{SA-MLP}^{KD}$ are improved over SA-MLPs by large margins and even outperform the teacher GNNs and the GLNN. The main reason for improvement is two-fold: 1) Analogous to GLNN, SA-MLP can learn graph knowledge from GNNs via cross-model KD. 2) Unlike GLNN, our SA-MLP encodes the structure in an alternative way, which may be complementary to GNNs, especially in the heterophily datasets (Squirrel and Arxiv-year).

\subsubsection{Inductive with Connection} As shown in Table~\ref{tab:node_weak_ind}, $\text{SA-MLP}^{KD}$ still consistently outperforms the others. We also make the following observations: 
1) The GLNN improves slightly on MLP. In the \textit{ind} scenario, distillation only occurs on the training nodes. Hence, the MLP without graph dependency generalizes limitedly on test nodes, especially for heterophily datasets where graph structure is essential. 
2) $\text{SA-MLP}^{KD}$ improves considerably after distillation. Since the SA-MLP can utilize the connection from training nodes, and as discussed in GLNN, the distillation from GNNs can further regularize students and enhance their generalization ability.

\subsubsection{Inductive without Connection} As shown in Table~\ref{tab:node_strong_ind}, we find the following: 1) For the without connection setting, an inconsistent structure distribution between training and test results in the poor performance of GNNs and SA-MLP, i.e., inferior to pure MLP in most datasets, which also causes the few improvement of GLNN as in the \text{ind w/c} setting.
2) The 1st-stage distillation can slightly improve the performance of SA-MLP, and the 2nd-stage further makes the SA-MLP to achieve the best performance. This phenomenon verifies the effectiveness of the latent structure embedding approximation with a two-stage distillation technique. It can enhance the model's generalization ability on unseen isolated nodes. We defer more discussion of the two-stage KD to Section~\ref{sec:mixed_ind}.

\begin{figure}[ht]
\begin{subfigure}[t]{.22\textwidth}
  \includegraphics[width=0.99\linewidth]{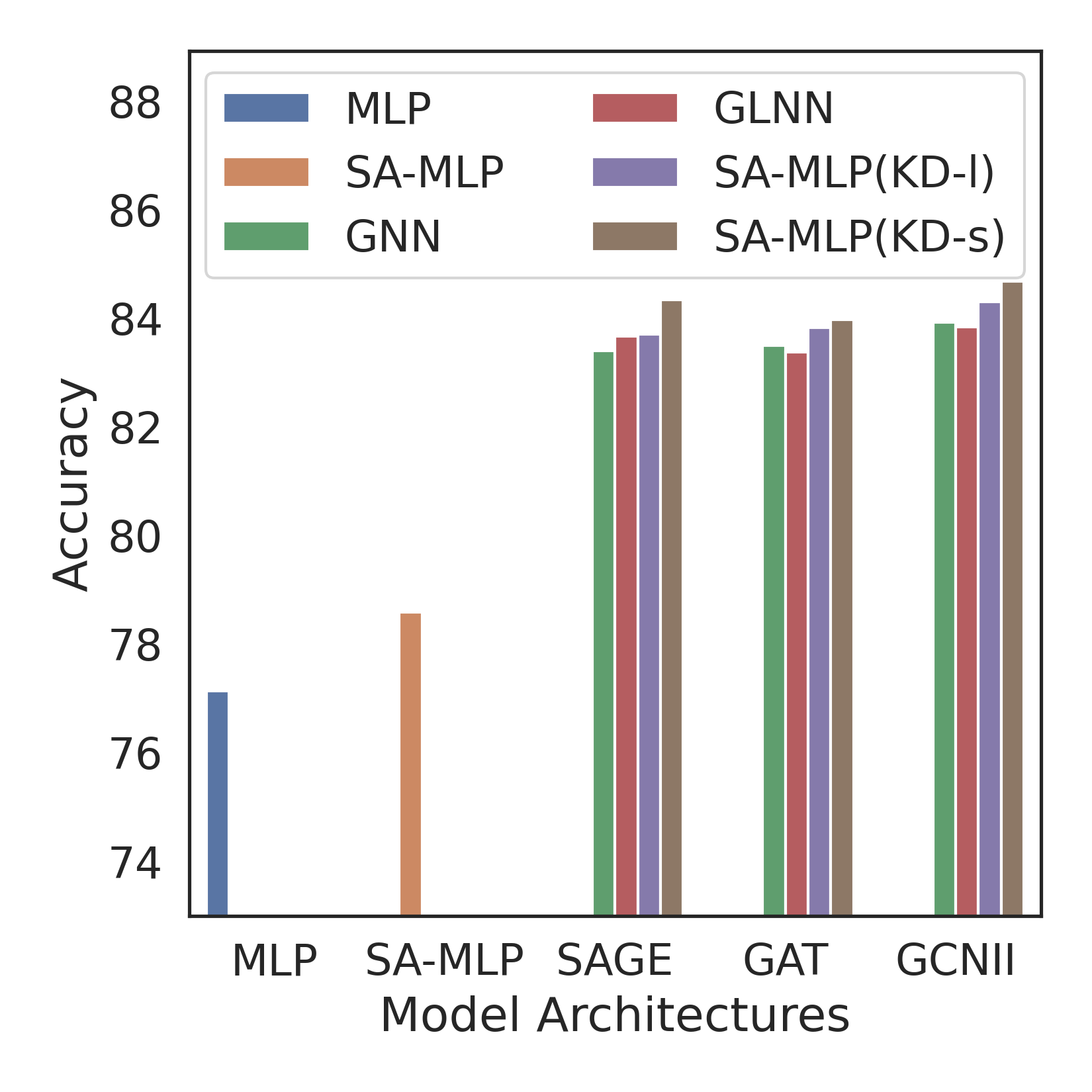}  
  \caption{\textit{trans}}
  \label{fig:abla-trans}
\end{subfigure}
\begin{subfigure}[t]{.22\textwidth}
  \includegraphics[width=0.99\linewidth]{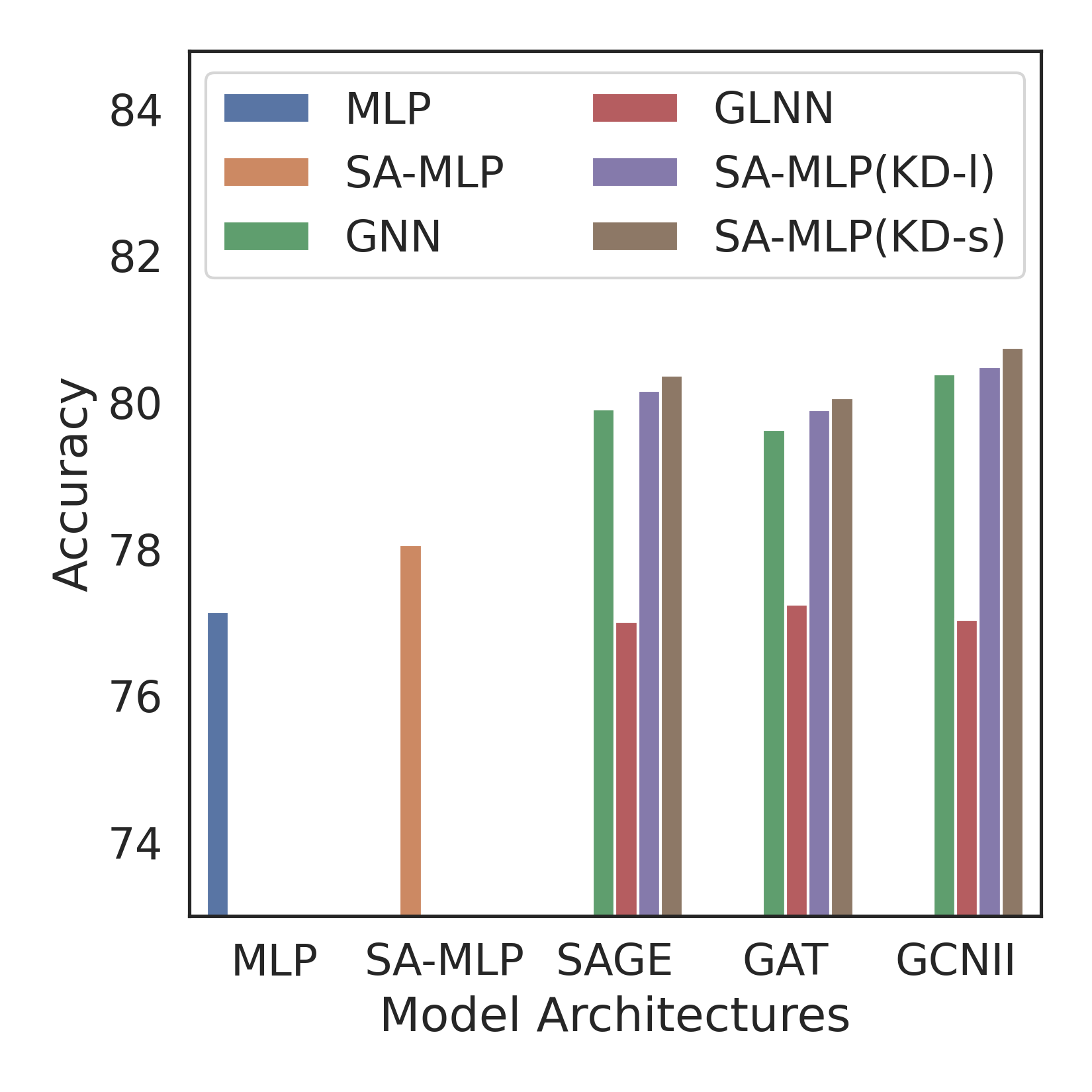}  
  \caption{\textit{ind w/c}}
  \label{fig:abla-ind}
\end{subfigure}
\caption{Mean accuracy over three citation datasets under different teacher architectures and distillation strategies. KD-l and KD-s indicate the standard logit-based KD and our structure-mixing KD, respectively.}
\label{fig:ablation}
\end{figure}

\subsection{In-Depth Analysis}
\subsubsection{Effects of Teacher GNN Architecture}We compare different teacher architectures, including SAGE~\cite{Hamilton2017InductiveRL}, GCNII~\cite{chen2020simple}, and GAT~\cite{GAT}. We choose three citation datasets in the \textit{trans} and \textit{ind w/c} settings and report the mean accuracy, since teacher GNNs generalize limitedly on \textit{ind w/o c} settings. Independent of the strategy of distillation, as in Figure~\ref{fig:ablation}, we see that $\text{SA-MLP}^{KD}$ can learn from different teachers and improve over SA-MLP, GLNN, and even GNN teachers on both \textit{trans} and \textit{ind w/c} settings. However, GLNN suffers from the weak generalization ability on \textit{ind w/c} among all teachers. The result shows that our SA-MLP can serve as a more general student than GLNN to accelerate the deployment of various GNNs.

\subsubsection{Effects of Structure-Mixing Distillation} We also study the effect of structure-mixing distillation under different teacher GNNs in the \textit{trans} and \textit{ind} settings. As Figure~\ref{fig:ablation} shows, the structure-mixing distillation strategy (KD-s) consistently outperforms the standard logit-based distillation (KD-l). These results show that the improvement of structure-mixing distillation is stable across GNN architectures and \textit{trans/ind} settings. 
In fact, a similar phenomenon has been observed in ~\cite{beyer2022knowledge} as well, i.e., \textit{mixup} can enhance the distillation performance in computer vision.

\subsubsection{Effects of two-stage KD in the Mixed Scenario}\label{sec:mixed_ind}

\begin{figure}[ht]
\begin{subfigure}[t]{.22\textwidth}
  \centering
    \includegraphics[width=0.99\textwidth]{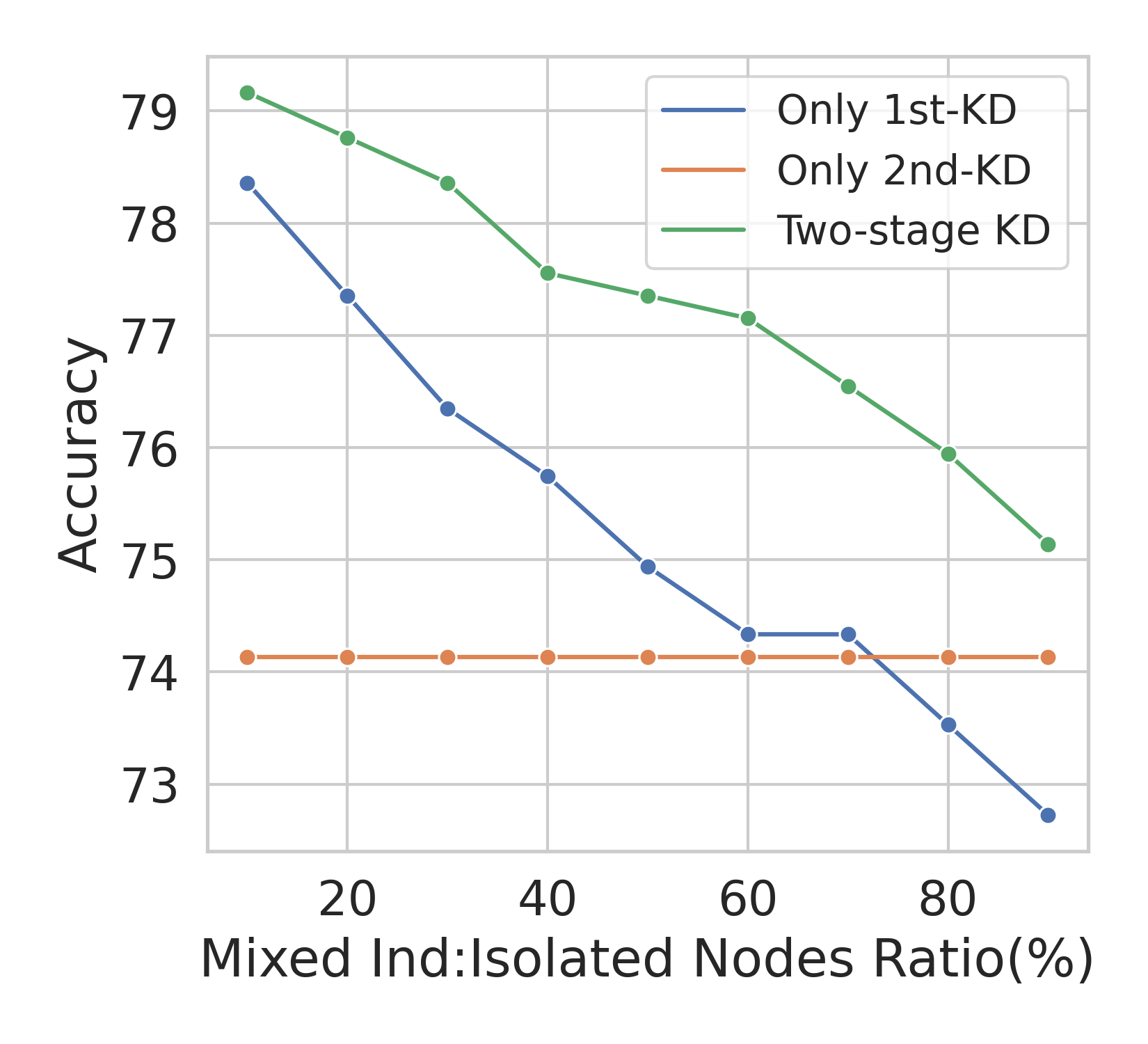}
  \caption{Cora (Homophily)}
\end{subfigure}
\begin{subfigure}[t]{.22\textwidth}
  \centering
    \includegraphics[width=0.99\textwidth]{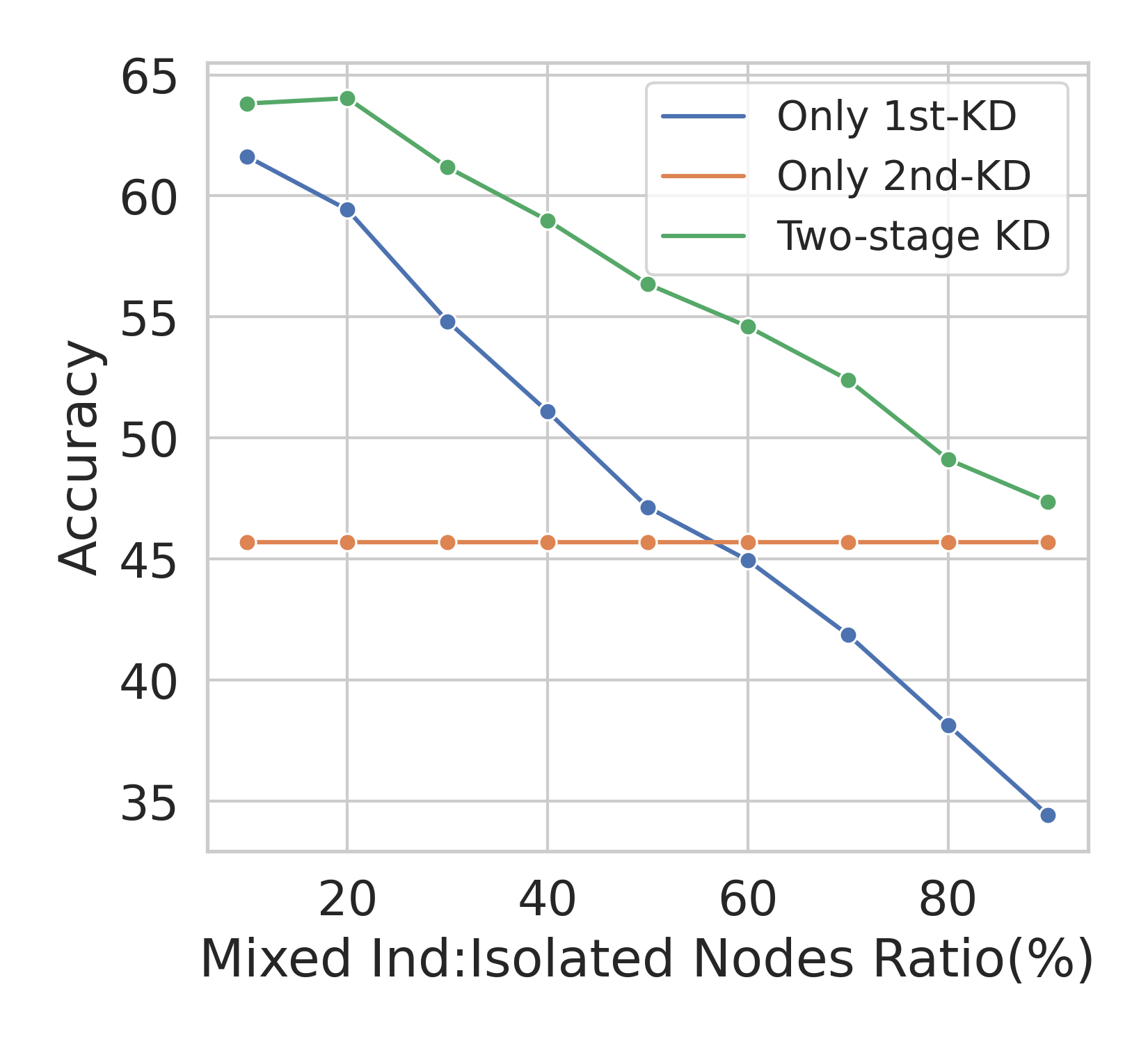}
  \caption{Chameleon (Heterophily)}
\end{subfigure}
\caption{Analysis of the two-stage distillation in the mixed \textit{ind}. The isolated node ratio indicates the mixed ratio of \textit{ind w/o c}.
}
\label{fig:mix}
\end{figure}

\begin{figure}[h]
\begin{subfigure}[t]{.22\textwidth}
  \centering
    \includegraphics[width=0.99\textwidth]{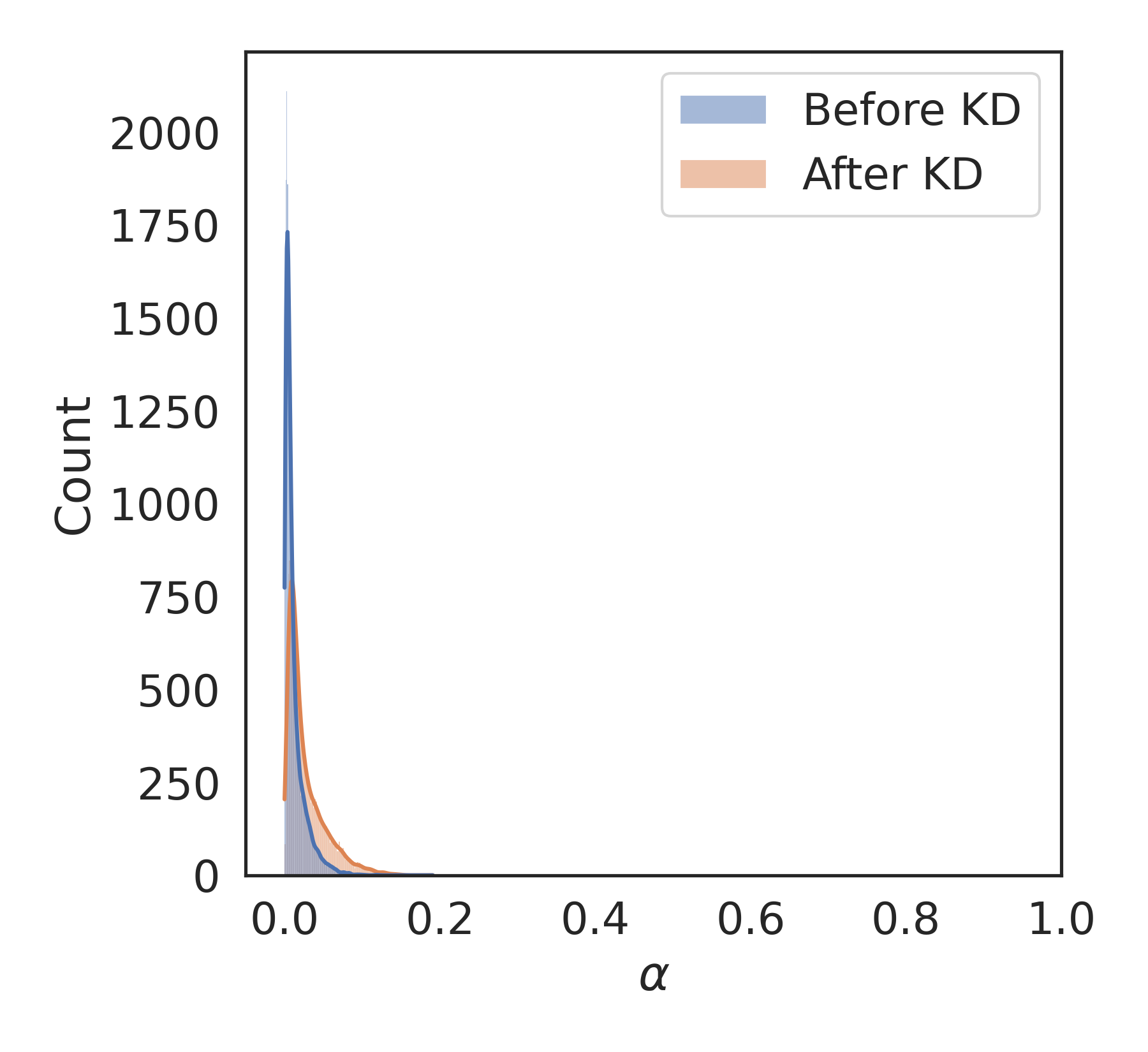}
  \caption{Pubmed (Homophily)}
\end{subfigure}
\begin{subfigure}[t]{.22\textwidth}
  \centering
    \includegraphics[width=0.99\textwidth]{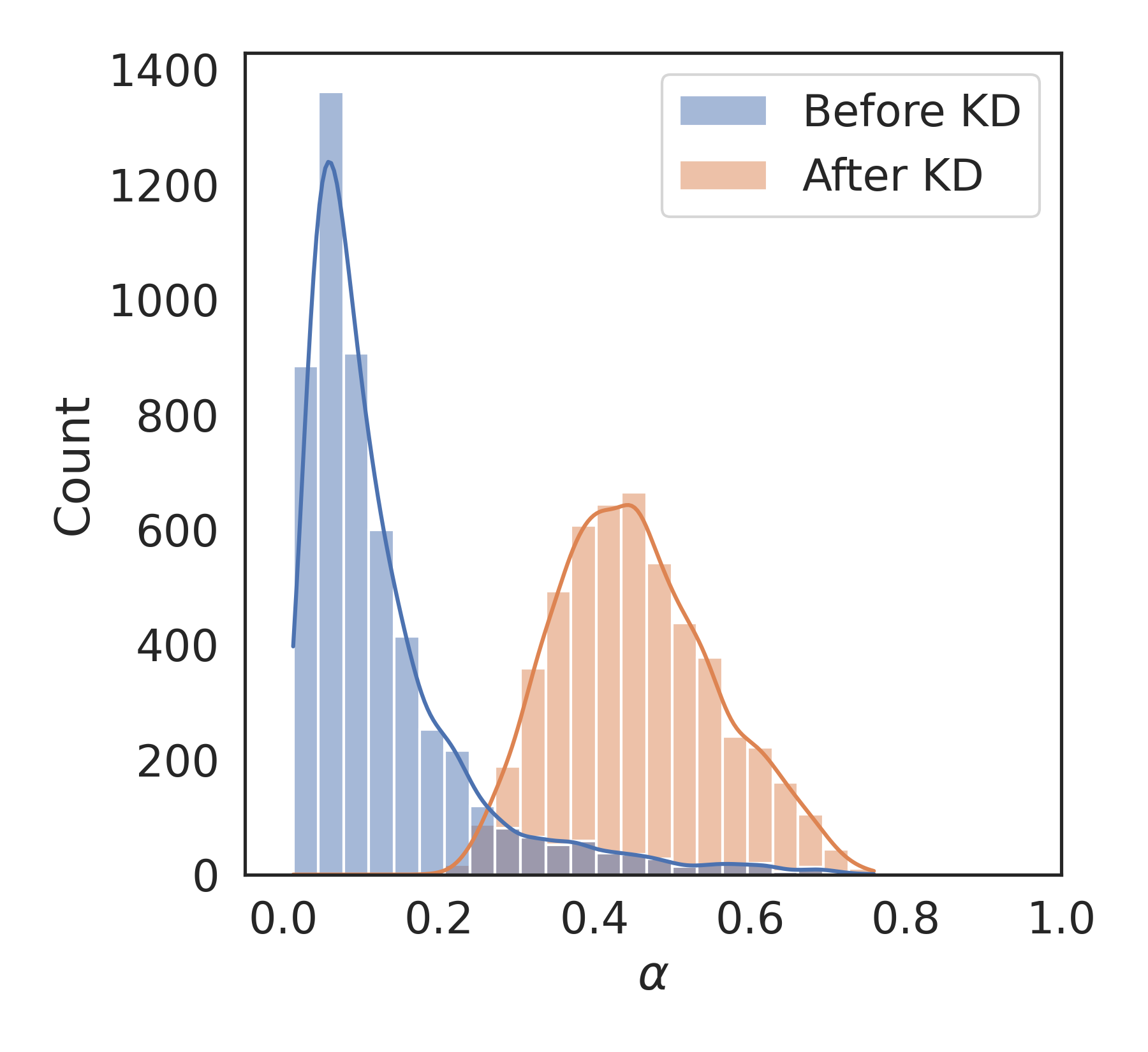}
  \caption{Squirrel (Heterophily)}
\end{subfigure}
\caption{The $\alpha$s' distribution before and after KD.}
\label{fig:alpha}
\end{figure}

To further show the superiority of the two-stage KD, we compare the performance of only the 1st-stage KD, only the 2nd-stage KD (using $\operatorname{MLP-2}_\mathbf{H_A}$ and finetune all modules), and the current two-stage KD in the mixed \textit{ind} scenario.
To construct the mixed \textit{ind} scenario, we randomly remove the connections of test nodes in the \textit{ind w/c} to enhance the isolated node ratio.
We can see that our two-stage method consistently outperforms all variants when increasing the mixing ratio. The reason is that it only finetunes the $\operatorname{MLP-2}_\mathbf{H_A}$ to deal with the isolated nodes and can still utilize structure knowledge learned from the 1st-stage KD for existing connections via $\operatorname{MLP}_\mathbf{H_A}$. However, other variants lack the compatibility, since they only focus on either the \textit{ind w/c} or \textit{ind w/o c}, leading to suboptimal results in mixed \textit{ind}. This 
indicates that our two-stage distillation is more appropriate for the mixed scenario that widely occurs in real-world online applications.

\subsubsection{Analysis of Interpretability}To study the interpretability of the learned SA-MLP, i.e., the contribution of features and structure, we visualize the distribution of $\alpha$ for each node before and after KD in Figure~\ref{fig:alpha}. Due to space limitations, we show the pattern of 
the typical Pubmed and Squirrel (heterophily) datasets, and the trends of other datasets are similar. 
For Pubmed, regardless of KD,
we can see that most $\alpha$s are close to 0. This phenomenon indicates that the structure provides less information, which can explain why the pure MLP can achieve almost 90\% test accuracy in Table~\ref{tab:node_trans}. In contrast, for the Squirrel, where structure information is essential for classification~\cite{zhu2021graph}, the $\alpha$s' distribution tends to shift into the middle after KD. The shifting indicates that GNN teaches SA-MLP to pay more attention to structure information and improve 
performance.

\subsubsection{Time Complexity} 
Following GLNN~\cite{zhang2021graph}, we show the speed comparison between GNNs with their inference acceleration and MLP-like students in the same experimental and GPU setting. The GNNs include the vanilla SAGE, quantized SAGE from FP32 to INT8 (QSAGE), SAGE with 50\% weights pruned (PSAGE), and the inference with neighbor sampling with fan-out 15 (NSSAGE).
As shown in Table~\ref{tab:exp-time}, all the MLP-like students achieve substantially faster inference than GNNs, since they discard the message-passing. 
$\text{SA-MLP}^{KD^2}$ is as fast as MLP, since it utilizes the latent structure embedding approximation without structure inputs.
Compared with GLNN, the SA-MLP needs to process the structure input. However, it is only slightly slower than GLNN due to the efficient implementation of sparse tensor multiplication in PyTorch~\cite{paszke2019pytorch}. 


\begin{table}[ht]
	\caption{ \textmd{Speed comparison between MLP-like students and other inference acceleration of SAGE. Numbers (in $ms$) are inductive inference times for 10 randomly chosen nodes on large-scale OGB Arxiv and Products. * indicates our implementation.}}
	\label{tab:exp-time}
	\begin{tabular}{l|c|rr}
		\toprule
		{Model} & Structure & {Arxiv} & {Products} \\
		\midrule
		SAGE  & \checkmark & 489.49   & 2071.30  \\
		QSAGE & \checkmark & 433.90 & 1946.49 \\
		PSAGE & \checkmark & 465.43 & 2001.46 \\
		NSSAGE &\checkmark & 91.03 & 107.31 \\
		\midrule
		GLNN & &  3.34 & 7.56 \\
		GLNN* & & 4.58 & 5.68 \\ 
		$\text{SA-MLP}^{KD^2}$ & & 7.14 & 8.89\\
		SA-MLP & \checkmark  & 10.64 & 14.81\\
		\bottomrule
	\end{tabular}
\end{table}

%% file: sections/Conclusion.tex
\section{{Conclusion}}
We have presented a message-passing free SA-MLP, a practical solution to address the deployment of GNNs via knowledge distillation. This is achieved by designing a simple yet effective structure-aware student MLP model and combining it with a novel structure-mixing knowledge distillation strategy. We also design a latent structure embedding approximation technique to deal with new nodes without connection in an inductive setting. Experiments on eight benchmark datasets show that SA-MLP enjoys as fast inference as MLP and as much accuracy as GNNs, while being explainable for its prediction to determine the contribution between features and structure. One future work is to investigate the application of SA-MLP to other downstream tasks, such as graph classification and link prediction on social networks.

%% file: sec-appendix.tex
\section{{Appendix}}
\subsection{Details of Datasets}
We provide the details of the five homophily datasets (connected nodes tend to be the same label) and three heterophily datasets (labels of connected nodes tend to be different) in the following:
\begin{itemize}
    \item {Homophily Datasets}
    \begin{itemize}
        \item \textit{Citeseer, Pubmed, Cora}~\cite{GCN}: For the basic citation datasets, nodes correspond to papers, edges correspond to citation links, the sparse bag-of-words are the feature representation of each node, and the label of each node represents the topic of the paper. 
        Note that, we use the public ten fully supervised data split(48\%/32\%/20\% for Train/Val/Test) in~\cite{pei2019geom,zhu2020beyond}. 
        Compared with the GLNN that uses only 20 nodes of each class for training, the results of our splits are more stable and reduce the possibility of overfitting.
        \item \textit{Arxiv}~\cite{hu2020open}: The Arxiv dataset is a large-scale citation network collected from all Computer Science ARXIV papers. Each node is an ARXIV paper, and edges are citation relations between papers. The features are 128-dimensional averaged word embeddings of each paper, and labels are subject areas of papers.
        \item \textit{Products}~\cite{hu2020open}: The Products dataset is a large-scale Amazon product co-purchasing network. Nodes represent products sold in Amazon, edges indicate the products purchased together, and features are 100-dimensional bag-of-words features.
    \end{itemize}
    \item {Heterophily Datasets}
    \begin{itemize}
        \item \textit{Squirrel, Chameleon}~\cite{pei2019geom}: Chameleon and Squirrel are web pages extracted from different topics in Wikipedia. Similar to WebKB, nodes and edges denote the web pages and hyperlinks among them, respectively, and informative nouns in the web pages are employed to construct the node features in the bag-of-word form. Webpages are labeled in terms of the average monthly traffic level.
        \item \textit{Arxiv-year}~\cite{lim2021large}: Modifying node labels of the Arxiv dataset to the year of paper, and the goal is to predict the year of paper publication that allows for evaluation of GNNs in large-scale non-homophilous settings.
    \end{itemize}
\end{itemize}
\subsection{Hyper-parameters Details}
We follow the number of layers setting of each model in GLNN and search other hyper-parameters, including hidden from [64, 128, 256], dropout from [0, 0.2, 0.5], learning rate (lr) from [0.01, 0.005, 0.05], weight decay (wd) from [0, 5e-4, 5e-5], $\delta$ from [0.2, 0.5], and $\lambda$ from [0.5, 0.8, 1] for distillation.

\subsection{Additional Comparison of GLNN+}
GLNN also provides a larger GLNN+ (scale hidden dimension from 256 to 1024 for Arxiv and 2048 for Product), with a larger capacity but a slower speed.
In this section, we provide additional experiments for the GLNN+ of the \textit{trans} and \textit{ind w/c} for large-scale OGB datasets. We omit other datasets since the performance of GLNN+ is similar to that of GLNN.
From Table~\ref{tab:glnn+}, we can find that the GLNN+ can improve the performance of large-scale OGB datasets under the \textit{trans} setting. However, it achieves similar results to GLNN under the \textit{ind} setting, which implies that the improvement of \textit{trans} for OGB datasets is due to the memory capacity, i.e., the larger parameters of GLNN+ can memorize all the teacher outputs. It still does not fully understand the structure information and generalizes limitedly on unseen test nodes under the \textit{ind} setting.
However, the improvement over both \textit{trans} and \textit{ind} of our SA-MLP is due to explicit structure awareness.


\begin{table}[]
\caption{Comparison with GLNN+}
    \label{tab:glnn+}

\begin{tabular}{l|l|llll}
\hline
Dataset & Setting & SAGE & GLNN & GLNN+ & SA-MLP \\ \hline
 & \textit{trans} & 70.92 & \cellcolor[HTML]{EFEFEF}63.46 & \cellcolor[HTML]{EFEFEF}\textbf{72.15} & 71.54 \\
\multirow{-2}{*}{Arxiv} & \textit{ind} & 67.69 & 56.35 & 56.56 & \textbf{68.01} \\ \hline
 & \textit{trans} & 78.61 & \cellcolor[HTML]{EFEFEF}68.86 & \cellcolor[HTML]{EFEFEF}77.65 & \textbf{79.02} \\
\multirow{-2}{*}{Product} & \textit{ind} & 65.55 & 62.45 & 62.58 & \textbf{67.46} \\ \hline
 & \textit{trans} & 51.85 & \cellcolor[HTML]{EFEFEF}46.22 & \cellcolor[HTML]{EFEFEF}51.02 & \textbf{53.31} \\
\multirow{-2}{*}{Arixv-year} & \textit{ind} & 48.42 & 36.92 & 36.81 & \textbf{49.55} \\ \hline
\end{tabular}
\end{table}

%% file: main.bbl

\begin{thebibliography}{49}


\ifx \showCODEN    \undefined \def \showCODEN     #1{\unskip}     \fi
\ifx \showDOI      \undefined \def \showDOI       #1{#1}\fi
\ifx \showISBNx    \undefined \def \showISBNx     #1{\unskip}     \fi
\ifx \showISBNxiii \undefined \def \showISBNxiii  #1{\unskip}     \fi
\ifx \showISSN     \undefined \def \showISSN      #1{\unskip}     \fi
\ifx \showLCCN     \undefined \def \showLCCN      #1{\unskip}     \fi
\ifx \shownote     \undefined \def \shownote      #1{#1}          \fi
\ifx \showarticletitle \undefined \def \showarticletitle #1{#1}   \fi
\ifx \showURL      \undefined \def \showURL       {\relax}        \fi
\providecommand\bibfield[2]{#2}
\providecommand\bibinfo[2]{#2}
\providecommand\natexlab[1]{#1}
\providecommand\showeprint[2][]{arXiv:#2}

\bibitem[Ba et~al\mbox{.}(2016)]%
        {Ba2016LayerN}
\bibfield{author}{\bibinfo{person}{Jimmy Ba}, \bibinfo{person}{Jamie~Ryan
  Kiros}, {and} \bibinfo{person}{Geoffrey~E. Hinton}.}
  \bibinfo{year}{2016}\natexlab{}.
\newblock \showarticletitle{Layer normalization}.
\newblock \bibinfo{journal}{\emph{preprint arXiv:1607.06450}}
  (\bibinfo{year}{2016}).
\newblock


\bibitem[Bahri et~al\mbox{.}(2021)]%
        {bahri2021binary}
\bibfield{author}{\bibinfo{person}{Mehdi Bahri}, \bibinfo{person}{Ga{\'e}tan
  Bahl}, {and} \bibinfo{person}{Stefanos Zafeiriou}.}
  \bibinfo{year}{2021}\natexlab{}.
\newblock \showarticletitle{Binary graph neural networks}. In
  \bibinfo{booktitle}{\emph{Proceedings of the IEEE/CVF Conference on Computer
  Vision and Pattern Recognition}}. \bibinfo{pages}{9492--9501}.
\newblock


\bibitem[Barkan and Koenigstein(2016)]%
        {barkan2016item2vec}
\bibfield{author}{\bibinfo{person}{Oren Barkan} {and} \bibinfo{person}{Noam
  Koenigstein}.} \bibinfo{year}{2016}\natexlab{}.
\newblock \showarticletitle{Item2vec: neural item embedding for collaborative
  filtering}. In \bibinfo{booktitle}{\emph{2016 IEEE 26th International
  Workshop on Machine Learning for Signal Processing (MLSP)}}. IEEE,
  \bibinfo{pages}{1--6}.
\newblock


\bibitem[Beyer et~al\mbox{.}(2022)]%
        {beyer2022knowledge}
\bibfield{author}{\bibinfo{person}{Lucas Beyer}, \bibinfo{person}{Xiaohua
  Zhai}, \bibinfo{person}{Am{\'e}lie Royer}, \bibinfo{person}{Larisa Markeeva},
  \bibinfo{person}{Rohan Anil}, {and} \bibinfo{person}{Alexander Kolesnikov}.}
  \bibinfo{year}{2022}\natexlab{}.
\newblock \showarticletitle{Knowledge distillation: A good teacher is patient
  and consistent}. In \bibinfo{booktitle}{\emph{Proceedings of the IEEE/CVF
  Conference on Computer Vision and Pattern Recognition}}.
  \bibinfo{pages}{10925--10934}.
\newblock


\bibitem[Bo et~al\mbox{.}(2021)]%
        {bo2021beyond}
\bibfield{author}{\bibinfo{person}{Deyu Bo}, \bibinfo{person}{Xiao Wang},
  \bibinfo{person}{Chuan Shi}, {and} \bibinfo{person}{Huawei Shen}.}
  \bibinfo{year}{2021}\natexlab{}.
\newblock \showarticletitle{Beyond low-frequency information in graph
  convolutional networks}. In \bibinfo{booktitle}{\emph{Proceedings of the AAAI
  Conference on Artificial Intelligence}}, Vol.~\bibinfo{volume}{35}.
  \bibinfo{pages}{3950--3957}.
\newblock


\bibitem[Chen et~al\mbox{.}(2018)]%
        {chen2018fastgcn}
\bibfield{author}{\bibinfo{person}{Jie Chen}, \bibinfo{person}{Tengfei Ma},
  {and} \bibinfo{person}{Cao Xiao}.} \bibinfo{year}{2018}\natexlab{}.
\newblock \showarticletitle{FastGCN: Fast learning with graph convolutional
  networks via importance sampling}. In \bibinfo{booktitle}{\emph{6th
  International Conference on Learning Representations, {ICLR} 2018}}.
\newblock


\bibitem[Chen et~al\mbox{.}(2020)]%
        {chen2020simple}
\bibfield{author}{\bibinfo{person}{Ming Chen}, \bibinfo{person}{Zhewei Wei},
  \bibinfo{person}{Zengfeng Huang}, \bibinfo{person}{Bolin Ding}, {and}
  \bibinfo{person}{Yaliang Li}.} \bibinfo{year}{2020}\natexlab{}.
\newblock \showarticletitle{Simple and deep graph convolutional networks}. In
  \bibinfo{booktitle}{\emph{International Conference on Machine Learning}}.
  PMLR, \bibinfo{pages}{1725--1735}.
\newblock


\bibitem[Deng and Zhang(2021)]%
        {Deng2021GraphFreeKD}
\bibfield{author}{\bibinfo{person}{Xiang Deng} {and} \bibinfo{person}{Zhongfei
  Zhang}.} \bibinfo{year}{2021}\natexlab{}.
\newblock \showarticletitle{Graph-Free Knowledge Distillation for Graph Neural
  Networks}. In \bibinfo{booktitle}{\emph{IJCAI}}.
\newblock


\bibitem[Gilmer et~al\mbox{.}(2017)]%
        {gilmer2017neural}
\bibfield{author}{\bibinfo{person}{Justin Gilmer}, \bibinfo{person}{Samuel~S
  Schoenholz}, \bibinfo{person}{Patrick~F Riley}, \bibinfo{person}{Oriol
  Vinyals}, {and} \bibinfo{person}{George~E Dahl}.}
  \bibinfo{year}{2017}\natexlab{}.
\newblock \showarticletitle{Neural message passing for quantum chemistry}. In
  \bibinfo{booktitle}{\emph{International Conference on Machine Learning}}.
  JMLR. org, \bibinfo{pages}{1263--1272}.
\newblock


\bibitem[Hamilton et~al\mbox{.}(2017)]%
        {Hamilton2017InductiveRL}
\bibfield{author}{\bibinfo{person}{William~L Hamilton}, \bibinfo{person}{Rex
  Ying}, {and} \bibinfo{person}{Jure Leskovec}.}
  \bibinfo{year}{2017}\natexlab{}.
\newblock \showarticletitle{Inductive representation learning on large graphs}.
  In \bibinfo{booktitle}{\emph{Advances in Neural Information Processing
  Systems}}. \bibinfo{pages}{1025--1035}.
\newblock


\bibitem[Han et~al\mbox{.}(2022)]%
        {Han2022GMixupGD}
\bibfield{author}{\bibinfo{person}{Xiaotian Han}, \bibinfo{person}{Zhimeng
  Jiang}, \bibinfo{person}{Ninghao Liu}, {and} \bibinfo{person}{Xia Hu}.}
  \bibinfo{year}{2022}\natexlab{}.
\newblock \showarticletitle{G-Mixup: Graph Data Augmentation for Graph
  Classification}. In \bibinfo{booktitle}{\emph{ICML}}.
\newblock


\bibitem[He et~al\mbox{.}(2017)]%
        {he2017neural}
\bibfield{author}{\bibinfo{person}{Xiangnan He}, \bibinfo{person}{Lizi Liao},
  \bibinfo{person}{Hanwang Zhang}, \bibinfo{person}{Liqiang Nie},
  \bibinfo{person}{Xia Hu}, {and} \bibinfo{person}{Tat-Seng Chua}.}
  \bibinfo{year}{2017}\natexlab{}.
\newblock \showarticletitle{Neural collaborative filtering}. In
  \bibinfo{booktitle}{\emph{Proceedings of the 26th International Conference on
  World Wide Web}}. \bibinfo{pages}{173--182}.
\newblock


\bibitem[Hinton et~al\mbox{.}(2015)]%
        {hinton2015distilling}
\bibfield{author}{\bibinfo{person}{Geoffrey Hinton}, \bibinfo{person}{Oriol
  Vinyals}, \bibinfo{person}{Jeff Dean}, {et~al\mbox{.}}}
  \bibinfo{year}{2015}\natexlab{}.
\newblock \showarticletitle{Distilling the knowledge in a neural network}.
\newblock \bibinfo{journal}{\emph{preprint arXiv:1503.02531}}
  \bibinfo{volume}{2}, \bibinfo{number}{7} (\bibinfo{year}{2015}).
\newblock


\bibitem[Hornik et~al\mbox{.}(1989)]%
        {hornik1989multilayer}
\bibfield{author}{\bibinfo{person}{Kurt Hornik}, \bibinfo{person}{Maxwell
  Stinchcombe}, {and} \bibinfo{person}{Halbert White}.}
  \bibinfo{year}{1989}\natexlab{}.
\newblock \showarticletitle{Multilayer feedforward networks are universal
  approximators}.
\newblock \bibinfo{journal}{\emph{Neural networks}} \bibinfo{volume}{2},
  \bibinfo{number}{5} (\bibinfo{year}{1989}), \bibinfo{pages}{359--366}.
\newblock


\bibitem[Hu et~al\mbox{.}(2020)]%
        {hu2020open}
\bibfield{author}{\bibinfo{person}{Weihua Hu}, \bibinfo{person}{Matthias Fey},
  \bibinfo{person}{Marinka Zitnik}, \bibinfo{person}{Yuxiao Dong},
  \bibinfo{person}{Hongyu Ren}, \bibinfo{person}{Bowen Liu},
  \bibinfo{person}{Michele Catasta}, {and} \bibinfo{person}{Jure Leskovec}.}
  \bibinfo{year}{2020}\natexlab{}.
\newblock \showarticletitle{Open graph benchmark: Datasets for machine learning
  on graphs}. In \bibinfo{booktitle}{\emph{Advances in Neural Information
  Processing Systems}}.
\newblock


\bibitem[Hu et~al\mbox{.}(2021)]%
        {hu2021graph}
\bibfield{author}{\bibinfo{person}{Yang Hu}, \bibinfo{person}{Haoxuan You},
  \bibinfo{person}{Zhecan Wang}, \bibinfo{person}{Zhicheng Wang},
  \bibinfo{person}{Erjin Zhou}, {and} \bibinfo{person}{Yue Gao}.}
  \bibinfo{year}{2021}\natexlab{}.
\newblock \showarticletitle{Graph-MLP: node classification without message
  passing in graph}.
\newblock \bibinfo{journal}{\emph{preprint arXiv:2106.04051}}
  (\bibinfo{year}{2021}).
\newblock


\bibitem[Joshi et~al\mbox{.}(2021)]%
        {Joshi2021OnRK}
\bibfield{author}{\bibinfo{person}{Chaitanya~K. Joshi}, \bibinfo{person}{Fayao
  Liu}, \bibinfo{person}{Xu Xun}, \bibinfo{person}{Jie Lin}, {and}
  \bibinfo{person}{Chuan-Sheng Foo}.} \bibinfo{year}{2021}\natexlab{}.
\newblock \showarticletitle{On Representation Knowledge Distillation for Graph
  Neural Networks}.
\newblock \bibinfo{journal}{\emph{preprint arXiv:2111.04964}}
  \bibinfo{volume}{abs/2111.04964} (\bibinfo{year}{2021}).
\newblock


\bibitem[Joulin et~al\mbox{.}(2016)]%
        {joulin2016fasttext}
\bibfield{author}{\bibinfo{person}{Armand Joulin}, \bibinfo{person}{Edouard
  Grave}, \bibinfo{person}{Piotr Bojanowski}, \bibinfo{person}{Matthijs Douze},
  \bibinfo{person}{H{\'e}rve J{\'e}gou}, {and} \bibinfo{person}{Tomas
  Mikolov}.} \bibinfo{year}{2016}\natexlab{}.
\newblock \showarticletitle{FastText.zip: Compressing text classification
  models}.
\newblock \bibinfo{journal}{\emph{preprint arXiv:1612.03651}}
  (\bibinfo{year}{2016}).
\newblock


\bibitem[Kingma and Ba(2014)]%
        {kingma2014adam}
\bibfield{author}{\bibinfo{person}{Diederik~P Kingma} {and}
  \bibinfo{person}{Jimmy Ba}.} \bibinfo{year}{2014}\natexlab{}.
\newblock \showarticletitle{Adam: A method for stochastic optimization}.
\newblock \bibinfo{journal}{\emph{preprint arXiv:1412.6980}}
  (\bibinfo{year}{2014}).
\newblock


\bibitem[Kipf and Welling(2017)]%
        {GCN}
\bibfield{author}{\bibinfo{person}{Thomas~N. Kipf} {and} \bibinfo{person}{Max
  Welling}.} \bibinfo{year}{2017}\natexlab{}.
\newblock \showarticletitle{Semi-supervised classification with graph
  convolutional networks}. In \bibinfo{booktitle}{\emph{International
  Conference on Learning Representations}}.
\newblock


\bibitem[Kuramochi and Karypis(2005)]%
        {kuramochi2005finding}
\bibfield{author}{\bibinfo{person}{Michihiro Kuramochi} {and}
  \bibinfo{person}{George Karypis}.} \bibinfo{year}{2005}\natexlab{}.
\newblock \showarticletitle{Finding frequent patterns in a large sparse graph}.
\newblock \bibinfo{journal}{\emph{Data mining and knowledge discovery}}
  \bibinfo{volume}{11}, \bibinfo{number}{3} (\bibinfo{year}{2005}),
  \bibinfo{pages}{243--271}.
\newblock


\bibitem[Lim et~al\mbox{.}(2021)]%
        {lim2021large}
\bibfield{author}{\bibinfo{person}{Derek Lim}, \bibinfo{person}{Felix Hohne},
  \bibinfo{person}{Xiuyu Li}, \bibinfo{person}{Sijia~Linda Huang},
  \bibinfo{person}{Vaishnavi Gupta}, \bibinfo{person}{Omkar Bhalerao}, {and}
  \bibinfo{person}{Ser~Nam Lim}.} \bibinfo{year}{2021}\natexlab{}.
\newblock \showarticletitle{Large scale learning on non-Homophilous graphs: New
  benchmarks and strong simple methods}.
\newblock \bibinfo{journal}{\emph{Advances in Neural Information Processing
  Systems}}  \bibinfo{volume}{34} (\bibinfo{year}{2021}).
\newblock


\bibitem[Liu et~al\mbox{.}(2021)]%
        {Liu2021PayAT}
\bibfield{author}{\bibinfo{person}{Hanxiao Liu}, \bibinfo{person}{Zihang Dai},
  \bibinfo{person}{David~R. So}, {and} \bibinfo{person}{Quoc~V. Le}.}
  \bibinfo{year}{2021}\natexlab{}.
\newblock \showarticletitle{Pay Attention to MLPs}. In
  \bibinfo{booktitle}{\emph{NeurIPS}}.
\newblock


\bibitem[Liu et~al\mbox{.}(2022a)]%
        {liu2022we}
\bibfield{author}{\bibinfo{person}{Ruiyang Liu}, \bibinfo{person}{Yinghui Li},
  \bibinfo{person}{Linmi Tao}, \bibinfo{person}{Dun Liang}, {and}
  \bibinfo{person}{Hai-Tao Zheng}.} \bibinfo{year}{2022}\natexlab{a}.
\newblock \showarticletitle{Are we ready for a new paradigm shift? a survey on
  visual deep mlp}.
\newblock \bibinfo{journal}{\emph{Patterns}} \bibinfo{volume}{3},
  \bibinfo{number}{7} (\bibinfo{year}{2022}), \bibinfo{pages}{100520}.
\newblock


\bibitem[Liu et~al\mbox{.}(2022b)]%
        {liu2022towards}
\bibfield{author}{\bibinfo{person}{Yixin Liu}, \bibinfo{person}{Yu Zheng},
  \bibinfo{person}{Daokun Zhang}, \bibinfo{person}{Hongxu Chen},
  \bibinfo{person}{Hao Peng}, {and} \bibinfo{person}{Shirui Pan}.}
  \bibinfo{year}{2022}\natexlab{b}.
\newblock \showarticletitle{Towards unsupervised deep graph structure
  learning}. In \bibinfo{booktitle}{\emph{Proceedings of the ACM Web Conference
  2022}}. \bibinfo{pages}{1392--1403}.
\newblock


\bibitem[Melas-Kyriazi(2021)]%
        {melas2021you}
\bibfield{author}{\bibinfo{person}{Luke Melas-Kyriazi}.}
  \bibinfo{year}{2021}\natexlab{}.
\newblock \showarticletitle{Do you even need attention? a stack of feed-forward
  layers does surprisingly well on imagenet}.
\newblock \bibinfo{journal}{\emph{preprint arXiv:2105.02723}}
  (\bibinfo{year}{2021}).
\newblock


\bibitem[Paszke et~al\mbox{.}(2019)]%
        {paszke2019pytorch}
\bibfield{author}{\bibinfo{person}{Adam Paszke}, \bibinfo{person}{Sam Gross},
  \bibinfo{person}{Francisco Massa}, \bibinfo{person}{Adam Lerer},
  \bibinfo{person}{James Bradbury}, \bibinfo{person}{Gregory Chanan},
  \bibinfo{person}{Trevor Killeen}, \bibinfo{person}{Zeming Lin},
  \bibinfo{person}{Natalia Gimelshein}, \bibinfo{person}{Luca Antiga},
  {et~al\mbox{.}}} \bibinfo{year}{2019}\natexlab{}.
\newblock \showarticletitle{Pytorch: An imperative style, high-performance deep
  learning library}.
\newblock \bibinfo{journal}{\emph{Advances in Neural Information Processing
  Systems}}  \bibinfo{volume}{32} (\bibinfo{year}{2019}).
\newblock


\bibitem[Pei et~al\mbox{.}(2020)]%
        {pei2019geom}
\bibfield{author}{\bibinfo{person}{Hongbin Pei}, \bibinfo{person}{Bingzhe Wei},
  \bibinfo{person}{Kevin Chen-Chuan Chang}, \bibinfo{person}{Yu Lei}, {and}
  \bibinfo{person}{Bo Yang}.} \bibinfo{year}{2020}\natexlab{}.
\newblock \showarticletitle{Geom-GCN: Geometric graph convolutional networks}.
  In \bibinfo{booktitle}{\emph{International Conference on Learning
  Representations}}.
\newblock


\bibitem[Sankar et~al\mbox{.}(2021)]%
        {sankar2021socialnet}
\bibfield{author}{\bibinfo{person}{Aravind Sankar}, \bibinfo{person}{Yozen
  Liu}, \bibinfo{person}{Jun Yu}, {and} \bibinfo{person}{Neil Shah}.}
  \bibinfo{year}{2021}\natexlab{}.
\newblock \showarticletitle{Graph Neural Networks for Friend Ranking in
  Large-Scale Social Platforms}. In \bibinfo{booktitle}{\emph{Proceedings of
  the Web Conference 2021}}. \bibinfo{pages}{2535–2546}.
\newblock


\bibitem[Sen et~al\mbox{.}(2008)]%
        {sen2008collective}
\bibfield{author}{\bibinfo{person}{Prithviraj Sen}, \bibinfo{person}{Galileo
  Namata}, \bibinfo{person}{Mustafa Bilgic}, \bibinfo{person}{Lise Getoor},
  \bibinfo{person}{Brian Galligher}, {and} \bibinfo{person}{Tina Eliassi-Rad}.}
  \bibinfo{year}{2008}\natexlab{}.
\newblock \showarticletitle{Collective classification in network data}.
\newblock \bibinfo{journal}{\emph{AI magazine}} \bibinfo{volume}{29},
  \bibinfo{number}{3} (\bibinfo{year}{2008}), \bibinfo{pages}{93--93}.
\newblock


\bibitem[Szegedy et~al\mbox{.}(2013)]%
        {szegedy2013intriguing}
\bibfield{author}{\bibinfo{person}{Christian Szegedy},
  \bibinfo{person}{Wojciech Zaremba}, \bibinfo{person}{Ilya Sutskever},
  \bibinfo{person}{Joan Bruna}, \bibinfo{person}{Dumitru Erhan},
  \bibinfo{person}{Ian Goodfellow}, {and} \bibinfo{person}{Rob Fergus}.}
  \bibinfo{year}{2013}\natexlab{}.
\newblock \showarticletitle{Intriguing properties of neural networks}.
\newblock \bibinfo{journal}{\emph{preprint arXiv:1312.6199}}
  (\bibinfo{year}{2013}).
\newblock


\bibitem[Tailor et~al\mbox{.}(2021)]%
        {tailor2020degree}
\bibfield{author}{\bibinfo{person}{Shyam~Anil Tailor}, \bibinfo{person}{Javier
  Fernandez-Marques}, {and} \bibinfo{person}{Nicholas~Donald Lane}.}
  \bibinfo{year}{2021}\natexlab{}.
\newblock \showarticletitle{Degree-Quant: Quantization-Aware Training for Graph
  Neural Networks}. In \bibinfo{booktitle}{\emph{International Conference on
  Learning Representations}}.
\newblock


\bibitem[Veličković et~al\mbox{.}(2018)]%
        {GAT}
\bibfield{author}{\bibinfo{person}{Petar Veličković},
  \bibinfo{person}{Guillem Cucurull}, \bibinfo{person}{Arantxa Casanova},
  \bibinfo{person}{Adriana Romero}, \bibinfo{person}{Pietro Liò}, {and}
  \bibinfo{person}{Yoshua Bengio}.} \bibinfo{year}{2018}\natexlab{}.
\newblock \showarticletitle{Graph Attention Networks}. In
  \bibinfo{booktitle}{\emph{International Conference on Learning
  Representations}}.
\newblock


\bibitem[Wang et~al\mbox{.}(2019)]%
        {wang2019knowledge}
\bibfield{author}{\bibinfo{person}{Hongwei Wang}, \bibinfo{person}{Fuzheng
  Zhang}, \bibinfo{person}{Mengdi Zhang}, \bibinfo{person}{Jure Leskovec},
  \bibinfo{person}{Miao Zhao}, \bibinfo{person}{Wenjie Li}, {and}
  \bibinfo{person}{Zhongyuan Wang}.} \bibinfo{year}{2019}\natexlab{}.
\newblock \showarticletitle{Knowledge-aware graph neural networks with label
  smoothness regularization for recommender systems}. In
  \bibinfo{booktitle}{\emph{Proceedings of the 25th {ACM} {SIGKDD}
  International Conference on Knowledge Discovery {\&} Data Mining, {KDD} 2019,
  Anchorage, AK, USA, August 4-8, 2019}}. \bibinfo{publisher}{{ACM}},
  \bibinfo{pages}{968--977}.
\newblock
\urldef\tempurl%
\url{https://doi.org/10.1145/3292500.3330836}
\showDOI{\tempurl}


\bibitem[Wang et~al\mbox{.}(2020)]%
        {wang2020traffic}
\bibfield{author}{\bibinfo{person}{Xiaoyang Wang}, \bibinfo{person}{Yao Ma},
  \bibinfo{person}{Yiqi Wang}, \bibinfo{person}{Wei Jin}, \bibinfo{person}{Xin
  Wang}, \bibinfo{person}{Jiliang Tang}, \bibinfo{person}{Caiyan Jia}, {and}
  \bibinfo{person}{Jian Yu}.} \bibinfo{year}{2020}\natexlab{}.
\newblock \showarticletitle{Traffic flow prediction via spatial temporal graph
  neural network}. In \bibinfo{booktitle}{\emph{Proceedings of The Web
  Conference 2020}}. \bibinfo{pages}{1082--1092}.
\newblock


\bibitem[Wang et~al\mbox{.}(2021)]%
        {Wang2021MixupFN}
\bibfield{author}{\bibinfo{person}{Yiwei Wang}, \bibinfo{person}{Wei Wang},
  \bibinfo{person}{Yuxuan Liang}, \bibinfo{person}{Yujun Cai}, {and}
  \bibinfo{person}{Bryan Hooi}.} \bibinfo{year}{2021}\natexlab{}.
\newblock \showarticletitle{Mixup for Node and Graph Classification}.
\newblock \bibinfo{journal}{\emph{Proceedings of the Web Conference 2021}}
  (\bibinfo{year}{2021}).
\newblock


\bibitem[Yan et~al\mbox{.}(2020)]%
        {yan2020tinygnn}
\bibfield{author}{\bibinfo{person}{Bencheng Yan}, \bibinfo{person}{Chaokun
  Wang}, \bibinfo{person}{Gaoyang Guo}, {and} \bibinfo{person}{Yunkai Lou}.}
  \bibinfo{year}{2020}\natexlab{}.
\newblock \showarticletitle{Tinygnn: Learning efficient graph neural networks}.
  In \bibinfo{booktitle}{\emph{Proceedings of the 26th ACM SIGKDD International
  Conference on Knowledge Discovery \& Data Mining}}.
  \bibinfo{pages}{1848--1856}.
\newblock


\bibitem[Yang et~al\mbox{.}(2021)]%
        {yang2021extract}
\bibfield{author}{\bibinfo{person}{Cheng Yang}, \bibinfo{person}{Jiawei Liu},
  {and} \bibinfo{person}{Chuan Shi}.} \bibinfo{year}{2021}\natexlab{}.
\newblock \showarticletitle{Extract the knowledge of graph neural networks and
  go beyond it: An effective knowledge distillation framework}. In
  \bibinfo{booktitle}{\emph{Proceedings of the Web Conference 2021}}.
  \bibinfo{pages}{1227--1237}.
\newblock


\bibitem[Yang et~al\mbox{.}(2020)]%
        {yang2020distilling}
\bibfield{author}{\bibinfo{person}{Yiding Yang}, \bibinfo{person}{Jiayan Qiu},
  \bibinfo{person}{Mingli Song}, \bibinfo{person}{Dacheng Tao}, {and}
  \bibinfo{person}{Xinchao Wang}.} \bibinfo{year}{2020}\natexlab{}.
\newblock \showarticletitle{Distilling knowledge from graph convolutional
  networks}. In \bibinfo{booktitle}{\emph{Proceedings of the IEEE/CVF
  Conference on Computer Vision and Pattern Recognition}}.
  \bibinfo{pages}{7074--7083}.
\newblock


\bibitem[Zhang et~al\mbox{.}(2018)]%
        {zhang2018mixup}
\bibfield{author}{\bibinfo{person}{Hongyi Zhang}, \bibinfo{person}{Moustapha
  Cisse}, \bibinfo{person}{Yann~N Dauphin}, {and} \bibinfo{person}{David
  Lopez-Paz}.} \bibinfo{year}{2018}\natexlab{}.
\newblock \showarticletitle{mixup: Beyond Empirical Risk Minimization}. In
  \bibinfo{booktitle}{\emph{International Conference on Learning
  Representations}}.
\newblock


\bibitem[Zhang et~al\mbox{.}(2022)]%
        {zhang2021graph}
\bibfield{author}{\bibinfo{person}{Shichang Zhang}, \bibinfo{person}{Yozen
  Liu}, \bibinfo{person}{Yizhou Sun}, {and} \bibinfo{person}{Neil Shah}.}
  \bibinfo{year}{2022}\natexlab{}.
\newblock \showarticletitle{Graph-less Neural Networks: Teaching Old MLPs New
  Tricks Via Distillation}. In \bibinfo{booktitle}{\emph{International
  Conference on Learning Representations}}.
\newblock


\bibitem[Zhao et~al\mbox{.}(2020)]%
        {zhao2020learned}
\bibfield{author}{\bibinfo{person}{Yiren Zhao}, \bibinfo{person}{Duo Wang},
  \bibinfo{person}{Daniel Bates}, \bibinfo{person}{Robert Mullins},
  \bibinfo{person}{Mateja Jamnik}, {and} \bibinfo{person}{Pietro Lio}.}
  \bibinfo{year}{2020}\natexlab{}.
\newblock \showarticletitle{Learned low precision graph neural networks}.
\newblock \bibinfo{journal}{\emph{preprint arXiv:2009.09232}}
  (\bibinfo{year}{2020}).
\newblock


\bibitem[Zheng et~al\mbox{.}(2022)]%
        {Zheng2022ColdBD}
\bibfield{author}{\bibinfo{person}{Wenqing Zheng}, \bibinfo{person}{Edward~W.
  Huang}, \bibinfo{person}{Nikhil~S. Rao}, \bibinfo{person}{Sumeet Katariya},
  \bibinfo{person}{Zhangyang Wang}, {and} \bibinfo{person}{Karthik Subbian}.}
  \bibinfo{year}{2022}\natexlab{}.
\newblock \showarticletitle{Cold Brew: Distilling Graph Node Representations
  with Incomplete or Missing Neighborhoods}.
\newblock \bibinfo{journal}{\emph{preprint arXiv:2111.04840}}
  \bibinfo{volume}{abs/2111.04840} (\bibinfo{year}{2022}).
\newblock


\bibitem[Zhou et~al\mbox{.}(2021)]%
        {zhou2021accelerating}
\bibfield{author}{\bibinfo{person}{Hongkuan Zhou}, \bibinfo{person}{Ajitesh
  Srivastava}, \bibinfo{person}{Hanqing Zeng}, \bibinfo{person}{Rajgopal
  Kannan}, {and} \bibinfo{person}{Viktor Prasanna}.}
  \bibinfo{year}{2021}\natexlab{}.
\newblock \showarticletitle{Accelerating large scale real-time GNN inference
  using channel pruning}.
\newblock \bibinfo{journal}{\emph{Proceedings of the VLDB Endowment}}
  \bibinfo{volume}{14}, \bibinfo{number}{9} (\bibinfo{year}{2021}),
  \bibinfo{pages}{1597--1605}.
\newblock


\bibitem[Zhu et~al\mbox{.}(2021a)]%
        {zhu2021graph}
\bibfield{author}{\bibinfo{person}{Jiong Zhu}, \bibinfo{person}{Ryan~A Rossi},
  \bibinfo{person}{Anup Rao}, \bibinfo{person}{Tung Mai},
  \bibinfo{person}{Nedim Lipka}, \bibinfo{person}{Nesreen~K Ahmed}, {and}
  \bibinfo{person}{Danai Koutra}.} \bibinfo{year}{2021}\natexlab{a}.
\newblock \showarticletitle{Graph neural networks with heterophily}. In
  \bibinfo{booktitle}{\emph{Proceedings of the AAAI Conference on Artificial
  Intelligence}}, Vol.~\bibinfo{volume}{35}. \bibinfo{pages}{11168--11176}.
\newblock


\bibitem[Zhu et~al\mbox{.}(2020)]%
        {zhu2020beyond}
\bibfield{author}{\bibinfo{person}{Jiong Zhu}, \bibinfo{person}{Yujun Yan},
  \bibinfo{person}{Lingxiao Zhao}, \bibinfo{person}{Mark Heimann},
  \bibinfo{person}{Leman Akoglu}, {and} \bibinfo{person}{Danai Koutra}.}
  \bibinfo{year}{2020}\natexlab{}.
\newblock \showarticletitle{Beyond homophily in graph neural networks: Current
  limitations and effective designs}. In \bibinfo{booktitle}{\emph{Advances in
  Neural Information Processing Systems}}, Vol.~\bibinfo{volume}{33}.
\newblock


\bibitem[Zhu et~al\mbox{.}(2021b)]%
        {zhu2021deep}
\bibfield{author}{\bibinfo{person}{Yanqiao Zhu}, \bibinfo{person}{Weizhi Xu},
  \bibinfo{person}{Jinghao Zhang}, \bibinfo{person}{Qiang Liu},
  \bibinfo{person}{Shu Wu}, {and} \bibinfo{person}{Liang Wang}.}
  \bibinfo{year}{2021}\natexlab{b}.
\newblock \showarticletitle{Deep graph structure learning for robust
  representations: A survey}.
\newblock \bibinfo{journal}{\emph{preprint arXiv:2103.03036}}
  (\bibinfo{year}{2021}).
\newblock


\bibitem[Zhuang et~al\mbox{.}(2022)]%
        {Zhuang2022DataFreeAK}
\bibfield{author}{\bibinfo{person}{Yu-Lin Zhuang}, \bibinfo{person}{Lingjuan
  Lyu}, \bibinfo{person}{Chuan Shi}, \bibinfo{person}{Carl Yang}, {and}
  \bibinfo{person}{Lichao Sun}.} \bibinfo{year}{2022}\natexlab{}.
\newblock \showarticletitle{Data-Free Adversarial Knowledge Distillation for
  Graph Neural Networks}. In \bibinfo{booktitle}{\emph{IJCAI}}.
\newblock


\bibitem[Zou et~al\mbox{.}(2019)]%
        {zou2019layer}
\bibfield{author}{\bibinfo{person}{Difan Zou}, \bibinfo{person}{Ziniu Hu},
  \bibinfo{person}{Yewen Wang}, \bibinfo{person}{Song Jiang},
  \bibinfo{person}{Yizhou Sun}, {and} \bibinfo{person}{Quanquan Gu}.}
  \bibinfo{year}{2019}\natexlab{}.
\newblock \showarticletitle{Layer-dependent importance sampling for training
  deep and large graph convolutional networks}.
\newblock \bibinfo{journal}{\emph{Advances in neural information processing
  systems}}  \bibinfo{volume}{32} (\bibinfo{year}{2019}).
\newblock


\end{thebibliography}
